\documentclass[journal]{IEEEtran}
\usepackage{cite}
\usepackage{amsmath,amssymb,amsfonts}
\usepackage{amsthm}
\usepackage{algorithmic}
\usepackage{algorithm}
\usepackage{graphicx}
\usepackage{textcomp}
\usepackage{xcolor}
\usepackage{booktabs}
\usepackage{array}
\usepackage{subfigure}
\usepackage{multirow}
\usepackage{threeparttable} 
\usepackage{tablefootnote}
\usepackage[colorlinks,bookmarksopen,bookmarksnumbered,citecolor=green, linkcolor=red, urlcolor=green]{hyperref}

\theoremstyle{break}

\ifCLASSINFOpdf
\else
\fi
\begin{document}

\title{Vehicle Dynamics Embedded World Models for Autonomous Driving}

\author{Huiqian~Li, Wei~Pan, \textit{Member},  \textit{IEEE}, Haodong~Zhang, Jin~Huang, Zhihua~Zhong
	\thanks{Manuscript received, 2024; revised , 2024. \textit{(Corresponding Authors: Wei Pan, Jin Huang.)}}
	\thanks{This Research is sponsored in part by the NSFC Program (No. U20A20285,
		No. 52122217, No. 52221005), and is also supported by China Scholarship
		Council (CSC)}
	\thanks{H. Li, H. Zhang, J. Huang are with School of Vehicle and Mobility, Tsinghua University, Beijing 100084, China (e-mail:thu\_lhq@163.com; zhdzyw@163.com; huangjin@tsinghua.edu.cn)}
	\thanks{Wei Pan is with the Department of Computer Science, The University of
	Manchester, M13 9PL Manchester, U.K. (e-mail: wei.pan@manchester.ac.uk).}
	\thanks{Zhihua Zhong is with the Tsinghua University, Chinese Academy of Engineering, Beijing 100088, China.}}

\markboth{Journal of \LaTeX\ Class Files,~Vol.~14, No.~8, August~2015}%
{Shell \MakeLowercase{\textit{et al.}}: Bare Demo of IEEEtran.cls for IEEE Transactions on Magnetics Journals}

\IEEEtitleabstractindextext{%
\begin{abstract}
	World models have gained significant attention as a promising approach for autonomous driving. By emulating human-like perception and decision-making processes, these models can predict and adapt to dynamic environments. Existing methods typically map high-dimensional observations into compact latent spaces and learn optimal policies within these latent representations. However, prior work usually jointly learns ego-vehicle dynamics and environmental transition dynamics from the image input, leading to inefficiencies and a lack of robustness to variations in vehicle dynamics. To address these issues, we propose the Vehicle Dynamics embedded Dreamer (VDD) method, which decouples the modeling of ego-vehicle dynamics from environmental transition dynamics. This separation allows the world model to generalize effectively across vehicles with diverse parameters. Additionally, we introduce two strategies to further enhance the robustness of the learned policy: Policy Adjustment during Deployment (PAD) and Policy Augmentation during Training (PAT). Comprehensive experiments in simulated environments demonstrate that the proposed model significantly improves both driving performance and robustness to variations in vehicle dynamics, outperforming existing approaches.
\end{abstract}

}

\maketitle
\IEEEdisplaynontitleabstractindextext
\IEEEpeerreviewmaketitle

\section{Introduction}
\label{introduction}

\IEEEPARstart{A}{chieving} autonomous driving in complex traffic scenarios relies on a comprehensive understanding of the environment \cite{agro2023implicit}. Recent advancements in artificial intelligence have facilitated significant breakthroughs in autonomous driving algorithms. These algorithms are progressively transitioning from modular pipeline architectures \cite{thrun2006stanley} to end-to-end models \cite{hu2023planning}, enhancing their overall performance and adaptability. 
However, despite the impressive capabilities of learning-based end-to-end models, they continue to struggle with pattern recognition tasks that humans perform effortlessly. This discrepancy arises from humans' inherent ability to apply 'common sense' and intuitive reasoning about potential futures, forming the foundation of human interaction with the world \cite{keller2012sensorimotor}. 
To bridge this gap, world models have emerged as a promising solution, endowing systems with the ability to predict and adapt to dynamic environments by simulating human-like perception and decision-making processes \cite{guan2024world}. 
In the context of autonomous driving, world models leverage historical data to simulate the states and dynamics of the driving environment and predict the potential outcomes of the vehicle's actions. 
This paper explores the application of world models in the planning and decision-making tasks of autonomous vehicles.

\begin{figure}[h]
	\centering
	\includegraphics[width=0.5\textwidth]{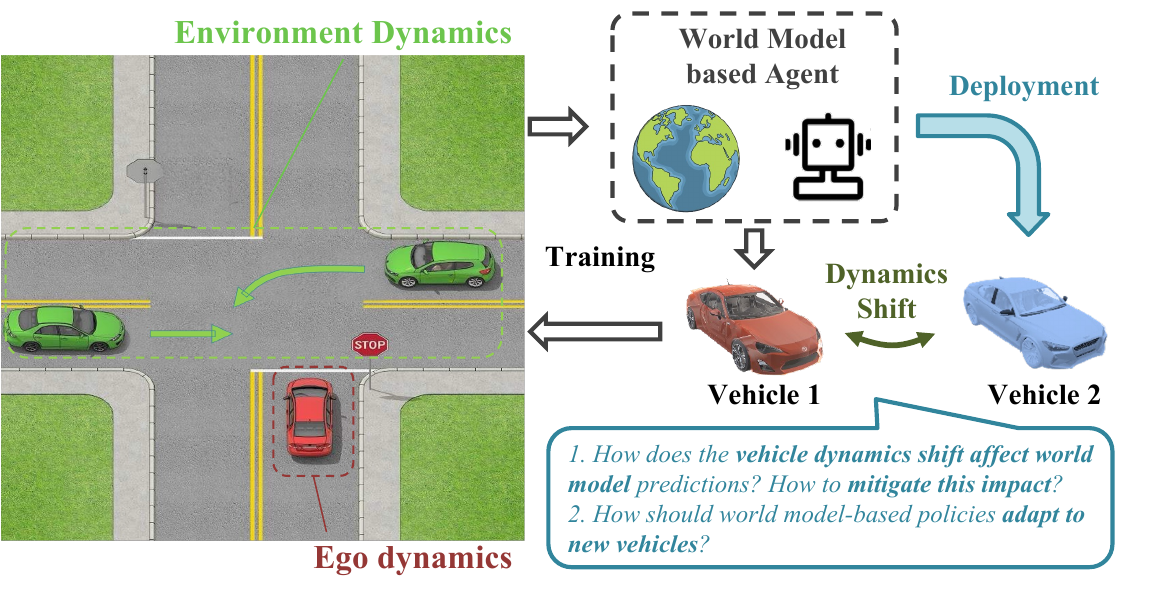}
	\caption{Illustration of the research motivation of this paper.}
	\label{fig:motivation}
\end{figure}

Learning an effective world model for vision-based control in autonomous driving remains a challenging endeavor \cite{pan2023model}. 
One primary reason is that the observation sequences collected by autonomous vehicles are high-dimensional and driven by multiple sources of physical dynamics, as shown on the left side of Fig.~\ref{fig:motivation}.
To address this challenge, existing studies have attempted to decouple the dynamics from different sources.
Gao et al. \cite{gao2024enhance} proposed a semantic masked world model that learns the transition dynamics of driving-relevant states through a semantic filter, thereby reducing interference from driving-irrelevant information in sensor inputs and improving the sample efficiency. 
Pan et al. \cite{pan2022iso} decomposed the spatio-temporal dynamics of the driving environment into controllable components that can respond to actions and uncontrollable components beyond the agent's control, enhancing the model's performance and robustness. 
However, these studies often focus solely on image inputs and output actions, overlooking the state information of the vehicle, such as speed, heading angle, yaw rate, and acceleration \cite{wang2023driving,li2024think2drive}. These low-dimensional dynamic state variables are readily accessible in autonomous vehicles and have been shown to be beneficial for planning tasks \cite{li2024ego}. 

Generalization in environments with dynamics shift constitutes another significant challenge for end-to-end autonomous driving based on world models \cite{chen2024end}. 
Current learning-based methods are based on the closed-world assumption, which assumes that the training and testing environments have the same distribution \cite{zhu2024open}. 
In the context of autonomous driving, real-world operational environments are often subject to frequent and unpredictable dynamics changes. For instance, an autonomous vehicle may need to navigate through urban areas with diverse weather and traffic conditions, or it might experience variations in its own attributes, such as changes in mass, as illustrated in Fig.~\ref{fig:motivation}.
Deploying policies trained in the nominal environment to settings with shifted dynamics settings can lead to substantial performance degradation. Nonetheless, limited research has addressed the implications of dynamics changes on world models \cite{ball2021augmented}. 
This study focuses on the influence of variations in ego vehicle dynamics on the efficacy of world model-based decision-making methods and proposes an approach to mitigate these effects.

To overcome the aforementioned challenges, this paper proposes a Vehicle Dynamics embedded Dreamer (VDD).
Main contributions of this paper are summarized as follows:
\begin{itemize}
	\item We present a world model that explicitly incorporates vehicle dynamics by decoupling the ego vehicle dynamics from the environment dynamics, thereby avoiding model mismatches when shifts in vehicle dynamics occur. Furthermore, we propose a behavior learning method for autonomous driving decision-making based on this decoupled world model.
	\item We introduce two strategies to mitigate the impact of changes in ego vehicle dynamics on decision-making behavior: Policy Adjustment during Deployment (PAD) and Policy Augmentation during Training (PAT). These strategies leverage the imagination capabilities of the proposed world model to generalize the policy to new environments with different ego vehicle dynamics parameters.
	\item We conduct extensive experiments on the proposed methods in simulated environments. The results demonstrate that merely adjusting control outputs is insufficient to accommodate drastic changes in ego vehicle dynamics. In contrast, the proposed VDD model significantly outperforms existing methods in terms of driving performance and robustness to variations in vehicle dynamics parameters.
\end{itemize}

The remainder of this paper is outlined as follows. Section \ref{related_work} discusses the related work on world models and distribution shifts problem for autonomous driving. Section \ref{preliminary} formulate the autonomous driving as a reinforcement learning (RL) problem and introduces the world models briefly. The methodology is presented in Section \ref{method} including the details of VDD and the strategies to improve robustness of policy. The experiment setups and results are described and discussed in Section \ref{experiments}. Section \ref{conclusion} concludes the paper.

\section{Related work}
\label{related_work} 
\subsection{World Models for Autonomous Driving}
World Models have emerged as a promising solution for autonomous driving on scenario generation and planning \cite{guan2024world}.
The application of world models in driving scenario generation facilitates the creation of diverse and realistic driving environments, significantly enriching the training datasets. This enrichment enhances the robustness of autonomous driving systems in handling rare and unforeseen scenarios. GAIA-1 trained a Transformer based world model using datasets collected from urban environments in UK, enabling it to learn and understand real-world traffic rules and key concepts in driving scenarios. It demonstrates the ability to comprehend and infer driving concepts not present in the training data, exhibiting counterfactual inference capabilities \cite{hu2023gaia}. 
In subsequent developments, world models evolved to incorporate multi-modal sensory inputs. For example, MUVO \cite{bogdoll2023muvo} and OccWorld \cite{zheng2023occworld} have successfully integrated LiDAR point clouds and 3D occupancy grids into world model inputs, significantly enhancing prediction accuracy and scenario generation quality. 
These works demonstrate the substantial capabilities of world models in interpreting and fusing complex sensory data, leading to improved understanding and modeling of driving environments. 
Furthermore, by leveraging scene understanding and prediction, world models have played a significant role in optimizing planning and control strategies in autonomous driving systems. MILE \cite{hu2022model} jointly learned environmental transition model and driving behaviors from offline datasets, significantly outperforming other models when tested in unseen towns and under novel weather conditions in the CARLA simulator. 
Think2Drive \cite{li2024think2drive} trained a compact latent world model to learn environmental dynamics, which was subsequently used as a neural simulator to train a planner. This method demonstrated a significant enhancement in the training efficiency of reinforcement learning and, for the first time, successfully solved all scenarios on the CARLA leaderboard v2. 
To improve the sampling efficiency of learning world models, SEM2 \cite{gao2024enhance} proposed a semantic masked world model that learns the transition dynamics of driving-relevant states through a semantic filter, thereby reducing interference from driving-irrelevant information in sensor inputs. ISODreamer++ \cite{pan2023model} optimized the inverse dynamics to encourage the world model to isolate the controllable state transitions from the mixed spatio-temporal changes in the environment, and optimize the policy based on the decoupled imagination. 
In summary, the advancement of world models is increasingly emphasizing the integration of more diverse input data and the improvement of sample efficiency.

\subsection{Autonomous Driving under 
Domain Shifts}

Domain shifts \cite{ben2010theory} are safety-critical challenges that autonomous vehicles inevitably face during real-world deployment \cite{sun2022shift}. Most related work focuses on distribution shift issues in perception tasks, which involve variations in weather conditions, illumination, and sensor characteristics \cite{li2023domain}. Some research addresses trajectory prediction and planning under distribution shifts, considering changes in geographic locations, traffic regulations, and social norms \cite{diehl2024lord,deng2021decision}.
Moreover, sim-to-real transfer presents a significant challenge, as it requires bridging the distributional gap between simulated environments and the real world. However, with regard to changes in vehicle dynamics, current research efforts have mainly concentrated on their impact on control systems, aiming to mitigate this impact by designing robust controllers \cite{li2022adaptive}. There has been little discussion of their influence on decision-making and planning. However, the following sections of this paper reveal that this influence deserves considerable attention in certain situations.

There has been a substantial amount of work studying how to detect and address different types of domain shifts \cite{filos2020can}. In the case that data in target domain is easily accessible, domain adaptation is a popular approach that can generalize model across the distribution shift \cite{so2022sim}. 
Sun et al. \cite{sun2022shift} introduced a large synthetic dataset SHIFT for autonomous driving perception tasks, which allows inexpensive data collection and annotation under discrete and continuous distribution shifts. This work demonstrated that conclusions drawn from the dataset hold in real-world datasets. 
Xing et al. \cite{xing2021domain} proposed a two-stage training strategy for zero-shot policy transfer, which first learns a latent unified state representation (LUSR) that is consistent across multiple domains, and then trains policy in the source domain. The cross-distribution consistency of LUSR makes the learned policy generalize to other target domains without extra training. The similar idea is also applied in \cite{zhang2023resimad}, which uses domain-invariant representations to achieve a zero-shot transfer ability.
Domain randomization \cite{tobin2017domain} is another effective technique to boost generalization. It randomizes the inputs like simulator images to prevent overfitting, improve adaptability to shifted conditions, and encourage extraction of generalizable features. 
The application of the world model sheds new light on the issue of detecting and mitigating distribution shifts \cite{zollicoffernovelty}. Ball et al. \cite{ball2021augmented} proposed to augment the dynamics model with simple transformations, which helps capture potential shifts in physical properties of the robot and learn more robust policies. Mu et al. \cite{mu2021model} presented a novel framework that enables the agent to learn policy from enriched diverse imagination derived from world models, improving sample efficiency and robustness. Context information has also been incorporated into the world model to improve its generalization ability to shifted dynamics \cite{lee2020context,far2024manyworlds}.

\section{Preliminary}
\label{preliminary}

\subsection{Reinforcement Learning for Autonomous Driving}
We model the visual control problem of autonomous vehicles as a partially observable Markov decision process (POMDP) with a tuple $(\mathcal{S}, \mathcal{A},\mathcal{T},\mathcal{R},\mathcal{O})$, where $\mathcal S$ is the state space, $\mathcal A$ is the action space, $\mathcal O$ is the observation space, $\mathcal R(s_t, a_t)$ is the reward function, and $\mathcal T(s_{t+1}|s_t,a_t)$ is the state-transition distribution. At each timestep $t\in [0:T]$, the agent takes an action $a_t\in\mathcal A$ to interact with the environment and receives an observation $o_t\in\mathcal O$ and a reward $r_t=\mathcal R(s_t, a_t)$ from the environment. The objective of the agent is to learn a policy that maximizes the expected cumulative discounted reward over time $\mathbb E_p[\sum_{t=0}^{T} \gamma^t r_t]$.  {In this setting, the agent cannot access the true state $s_t \in \mathcal S$, reward function $\mathcal R(s_t, a_t)$ and transition function $\mathcal T(s_{t+1}|s_t,a_t)$ of the environment. Instead, it can only estimate true state and approximate functions using accessible observations.}

\subsection{Reinforcement Learning under Vehicle Dynamics Shift}

To demonstrate the impact of vehicle dynamics shift, we consider the trajectory formulation of optimization objectives. Let $p^\pi(\cdot)$ represent the probability over trajectory $\mathcal H=\{\tau=(s_0,a_0,r_0,s_1,...,a_T,r_T)\}$ generated by executing policy $\pi$ under dynamics $p$, given by:
\begin{equation}
	p^\pi(\tau) = p(s_0) \prod_{t=0}^{T-1} \pi(a_t|s_t) p(s_{t+1}|s_t,a_t).
\end{equation}
The reinforcement learning objective are then formalized as:
\begin{equation}
	\mathcal J(\pi) = \mathbb E_{p^\pi}[R(\tau)],
\end{equation}
where $R(\tau) = \sum_{t=0}^{T} \gamma^t r_t$ represent the discounted return of the trajectory. We consider the case where the dynamics of the ego vehicle is changed, while the environment transitions remain the same. In this paper, we refer to the environment in which the agent collects training data as the 'original environment' and the environment where vehicle dynamics change as the 'shifted environment'. Following the assumptions in \cite{pan2023model,deng2021decision}, we divide the state into two parts, the state of the ego vehicle $s^i$ and the state of the environment $s^e$. We assume that the environment transition is influenced by the ego vehicle state, rather than directly by the action. Thus, the transition function can be written as:
\begin{equation}
	\begin{aligned}
		p(s_{t+1}|s_t,a_t) = & p(s^e_{t+1},s^i_{t+1}|s^e_t,s^i_t,a_t) \\		
		= & p(s^e_{t+1}|s^e_t,s^i_t)p(s^i_{t+1}|s^i_t,a_t).
	\end{aligned}
\end{equation}
Under this assumption, changing the ego vehicle dynamics does not affect the environment dynamics. Then we consider the situation where the dynamics of the ego vehicle is shifted from $p$ to $p_M$, the discounted returns applying the original policy $\pi$ can be written as:
\begin{equation}
	\mathcal J_M(\pi) = \mathbb E_{p^\pi_M}[R(\tau)] 
	= \mathbb E_{p^\pi}[\frac{p^\pi_M(\tau)}{p^\pi(\tau)}R(\tau)].
\end{equation}
We first analyze the influence of the dynamics shift on the performance of the original policy, and then we try to find a policy that can adapt to the new dynamics to improve the performance, which satisfies the following equation:
\begin{equation}
	\pi_M = \arg\max_{\pi'}\mathcal J_M(\pi').
\end{equation}

\subsection{World Models}
World models describe the internal models of an agent, which encodes how world states relate to a given sensory input, evolve and respond to the agent's actions. A compact internal representation allows the agent to perform a sample-efficient prediction of the present and future states of the world, and further enables efficient planning in a low-dimensional space \cite{taniguchi2023world}.
World models are trained based on the experiences collected by the agent through interacting with the environment, which include observations $o_{1:T}$, actions $a_{1:T}$, rewards $r_{1:T}$, and episode continuation flags $c_{1:T}$. During training, we sample fixed-length sequences of data from the experience replay buffer, compute their loss function, and update the model parameters using backpropagation. The components of the world model are jointly optimized based on maximum likelihood estimation (MLE), and the optimization objective is to maximize the log-likelihood for the reconstruction of the observation sequence, also known as the log-model evidence, which can be written as:
\begin{equation}
    \begin{aligned}
    	&\ln p(x_{1:T}|a_{1:T})= \ln \prod_{t=1}^T p(x_t|s_t)p(s_t|s_{t-1},a_{t-1}) \\
        & \geq \sum_{t=1}^{T} \left( \underset{q(s_t|o_{\leq t}, a_{<t})}{\mathbb{E}} [\ln p(x_t|s_t)] \right.\\
        &\left. - \underset{q(s_{t-1}|o_{\leq t-1}, a_{<t-1})}{\mathbb{E}}[\text{KL}[q(s_t|o_{\leq t}, a_{<t})||p(s_t|s_{t-1},a_{t-1})]] \right),
    \end{aligned}
    \label{eq:model_evidence}
\end{equation}
where $x_t \dot{=}  \left\{ o_t, r_t, c_t \right\}$, and $q(s_t|o_{\leq t}, a_{<t})$ is the variational distribution used to approximate the true posterior. Using importance weighting technique and Jensen's inequality, we can obtain the evidence lower bound (ELBO). The objective of training world models is to maximize the ELBO, or minimize the variational free energy (VFE) \cite{friston2016active}:
\begin{equation}
	\begin{aligned}
		\text{VFE}& = - \text{ELBO}\\ & = \mathbb E_{q_{\phi}(s_{1:T}|a_{1:T}, o_{1:T})}\left[ \sum_{t=1}^{T} -\ln p_\phi(o_t|s_t) -\ln p_{\phi}(r_t|s_t) \right. \\
		&\left.  -\ln p_{\phi}(c_t|s_t)+ \text{KL}[q_\phi(s_t|o_t)||p_\phi(s_t|s_{t-1},a_{t-1})] \right].
		\label{eq:elbo}
	\end{aligned}
\end{equation}

\subsection{Vehicle Dynamics Model}
A vehicle dynamics model is a mathematical framework designed to simulate and analyze a vehicle's motion and behavior in response to different control inputs. These models capture the complex interaction among a vehicle's physical properties. The kinematic bicycle model is suitable for motion planning purposes when lateral acceleration is constrained below $0.5\mu g$ \cite{polack2017kinematic}. The vehicle dynamics model employed in this study is defined as follows \cite{Althoff2017a}:
\begin{equation}
	\begin{aligned}
			\dot{x} &= v \cos{\psi} \\
			\dot{y} &= v \sin{\psi} \\
			\dot{\psi} &= \frac{v}{l_{wb}}\tan({f_{\text{sw}}(a_{\text{lat}})}) \\
			\dot{v} &= f_{\text{acc}}(a_{\text{long}},v) \\
	\end{aligned}
	\label{eq:vehicle_dynamics}
\end{equation}
where $x,y,\psi,v$ represent the vehicle's position, heading, velocity. $a_{\text{sw}}$ is the front wheel steering angle, $\l_{wb}$ is the distance between the front and rear axles, $a_{\text{long}}$ is the throttle or brake input, $f_{\text{acc}}$ is the mapping from the throttle or brake value to the acceleration, and $f_{\text{sw}}$ is the mapping from the steering angle to the front wheel angle. 

\section{methodology}
\label{method}
In this section, we begin by introducing the overall structure of the world model embedded with vehicle dynamics, followed by a description of how behaviors are learned based on this model. Additionally, we present two approaches to improve the robustness of the decision-making process in response to changes in vehicle dynamics.

\subsection{Vehicle Dynamics Embedded World Model}
\subsubsection{Model Structure}
\begin{figure}[htbp]
	\centering
	\begin{subfigure}
		\centering
		\includegraphics[width=0.4\textwidth]{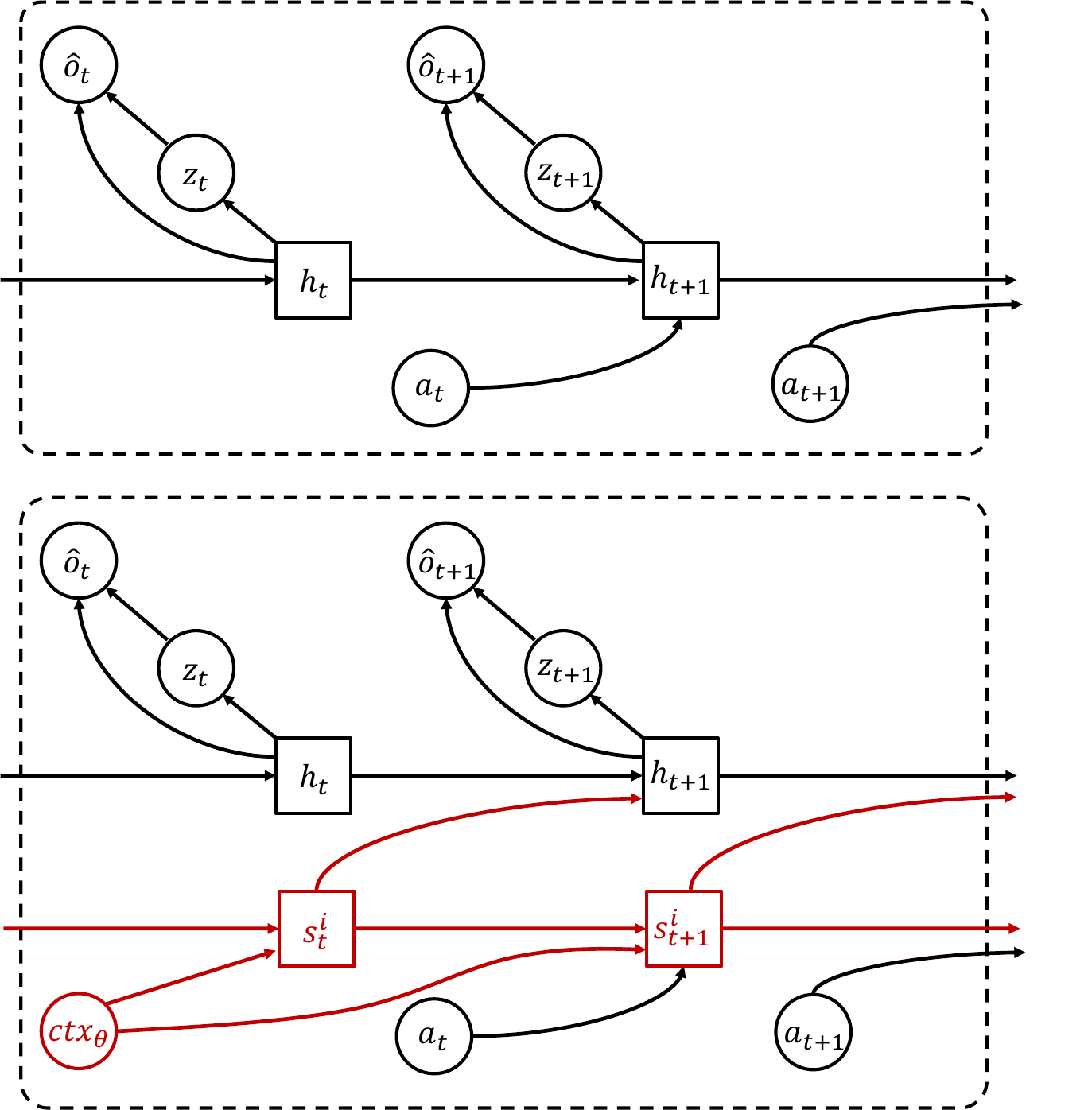}
	\end{subfigure}
	\caption{Recurrent state space model (top) and hierarchical context-aware recurrent state space model (bottom).}
	\label{fig:rssm}
\end{figure}

In this work, we adopt the Dreamer's framework \cite{hafner2020dreamer,hafner2021dreamerv2,hafner2023dreamerv3} to construct a world model for autonomous driving. 
Dreamer comprises three neural networks: the world model, the actor and the critic, each trained concurrently on replayed experience without gradient sharing. 
The internal state of the world model, known as the latent state, is modeled using a recurrent state space model (RSSM), as illustrated in Fig.~\ref{fig:rssm}. 
In the RSSM, the latent state is partitioned into a deterministic component $h_t$ and a stochastic component $z_t$, with future state predicted based on both pathways.
Assuming that agents generate control behaviors according to driving context \cite{ding2021epsilon}, we modify the RSSM into a hierarchical model where the deterministic pathway is controlled by the ego vehicle’s state. 
The ego vehicle dynamics are represented by a set of specific parameters, termed the vehicle dynamics context. Accordingly, we propose a hierarchical context-aware RSSM (hcRSSM), depicted in Fig.~\ref{fig:rssm}. In this hcRSSM, the lower level represents the vehicle dynamics model, influenced by the vehicle dynamics context, while the upper level captures how ego status affects the driving environment, referred to as the environmental dynamics model (EDM).

The learning process of the vehicle dynamics embedded world model is shown as Fig.~\ref{vdd-structure}. It consists of three main components: the environmental dynamics model, decoder heads and vehicle dynamics model. 
Initially, an encoder maps sensory inputs $o_t$ into stochastic representations $z_t$.
First, an encoder maps the sensory inputs $o_t$ to stochastic representations $z_t$. 
A sequential model with recurrent state $h_t$ then predicts the sequence of these representations based on the previous ego vehicle state $s^i_{t-1}$.
 {Since the reward function and episode continuation flag are not accessible directly, the combined representations $z_t$ and $h_t$ are fed into multiple decoder heads to predict rewards $\hat{r}_t$, episode continuation flags $\hat{c}_t \in \{0,1\}$ for computation of expected returns. Also, the model states are fed into the image decoders to reconstruct observations, ensuring that the representations are informative. }
The vehicle dynamics model predicts the current ego state $s^i_t$ based on the previous ego state $s^i_{t-1}$ and the action $a_t$. These model components are denoted as follows:
\begin{figure*}[h]
    \centering
    \begin{minipage}[b]{0.5\textwidth}
        \centering
        \includegraphics[width=\textwidth]{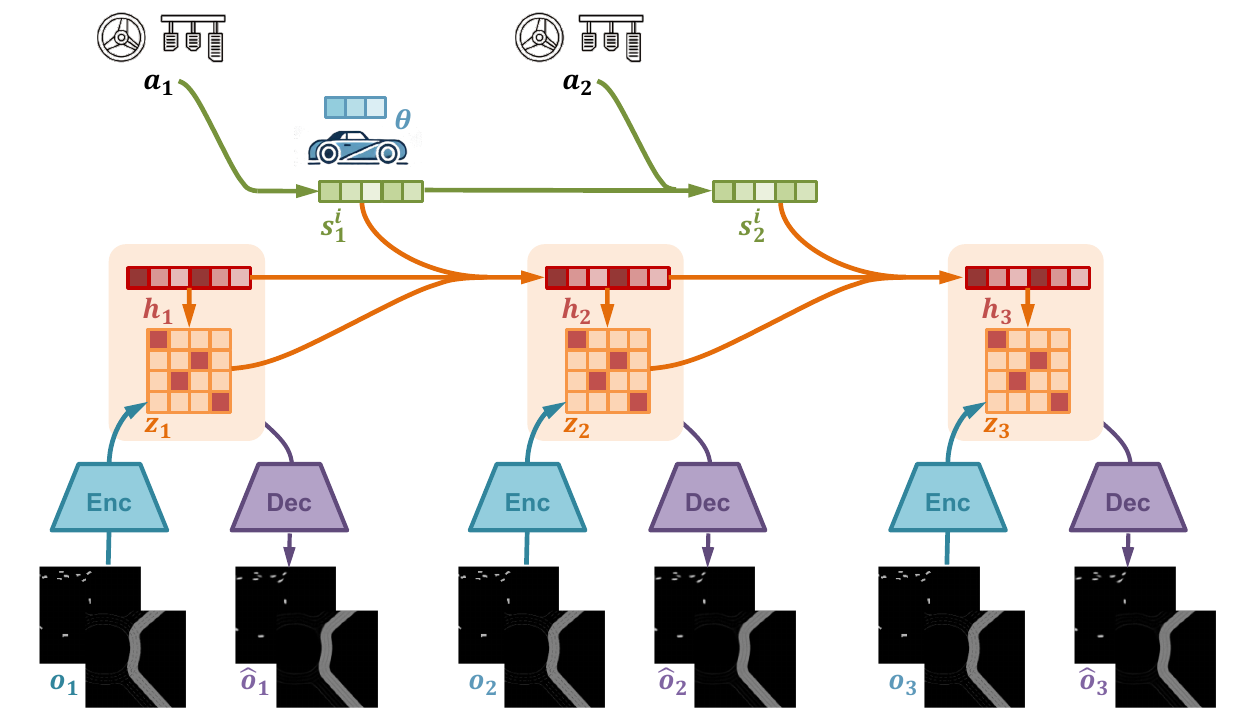}
    \end{minipage}
    \hspace{0.05\textwidth} 
    \begin{minipage}[b]{0.43\textwidth}
        \centering
        \includegraphics[width=\textwidth]{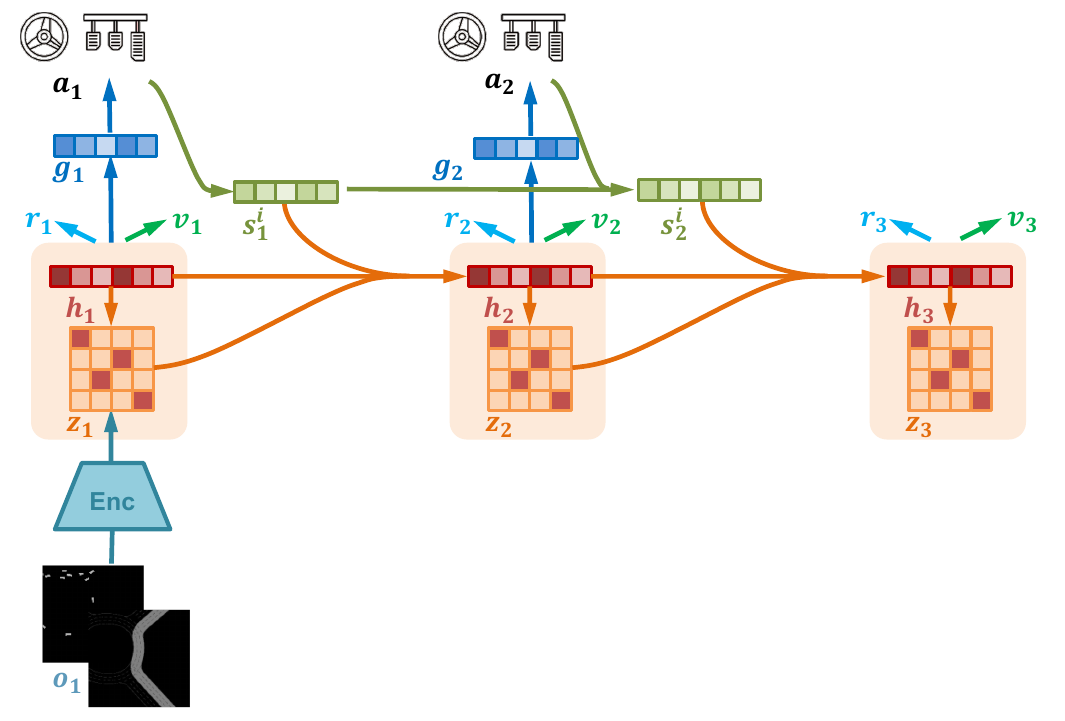}
    \end{minipage}
	\caption{The world model learning process (left) and behavior learning process (right) of the proposed VDD model. 
	The model receives bird’s-eye view (BEV) images as sensory inputs $o_t$, encoding them into discrete stochastic representations $z_t$. A sequential model with recurrent state $h_t$ then predicts the sequence of these representations based on the previous dynamics state of the ego vehicle, $s^i_{t-1}$. The actor and critic modules predict goals $g_t$ and values $v_t$, learning from trajectories of abstract representations generated by the world model. The controller of the ego vehicle subsequently produces the action $a_t$.}
	\label{vdd-structure}
\end{figure*}
\begin{equation}
	\begin{aligned}
		&\text{Sequential~model:} &h_t&=f_\phi(h_{t-1},z_{t-1},s^i_{t-1}) \\
		&\text{Encoder:}        &z_t&\sim q_\phi(z_t|h_t,o_t) \\
		&\text{Dynamics~predictor:} &\hat{z}_t&\sim p_\phi(z_t|h_t) \\
		&\text{Reward~predictor:} &\hat{r}_t&\sim p_\phi(r_t|z_t,h_t) \\
		&\text{Continuation~predictor:} &\hat{c}_t&\sim p_\phi(c_{t}|z_t,h_t) \\
		&\text{Decoder:}        &\hat{o}_t&\sim p_\phi(o_t|z_t,h_t) \\
		&\text{Vehicle~dynamics:} &s^i_t &= f_{\theta}(s^i_{t-1},a_{t})
	\end{aligned}
\end{equation}
Both the EDM and decoder heads are implemented as neural networks, with $\phi$ representing the model parameters. 
The vehicle dynamics model can be implemented either as a neural network, as in \cite{chen2021learning}, or as a parameterized physical model defined by \ref{eq:vehicle_dynamics} with parameter $\theta$. 
In this study, we implement the encoder model as convolutional neural networks (CNN) followed by a multi-layer perceptron (MLP) to compress image embeddings and recurrent state concatenation to stochastic latent representations. 
The sequential model is constructed as a recurrent neural network (RNN) with gated recurrent units (GRU) \cite{cho2014learning}. The dynamics predictor, continuation predictor and reward predictor are implemented as MLPs, while the observation decoder  uses a transposed CNN. 
The vehicle dynamics model receives ego state $s^i = [\Delta x, \Delta y, \Delta \psi,v,\dot{\psi}, \dot{v}]$ as inputs and employs a parameterized physical model to predict the subsequent ego state. Here, $\Delta x, \Delta y, \Delta \psi$ represent the relative position and orientation to the end waypoint of the current reference lane \cite{li2022metadrive}, while the definitions of $v, \dot{\psi}, \dot{v}$ align with those in Eq.~\ref{eq:vehicle_dynamics}.   { We observe that utilizing only the ego vehicle's acceleration $\dot v$ and angular velocity $\dot \psi$ as state inputs of sequential model yields better performance compared to employing the full state representation.}

\subsubsection{Loss Function}
The vehicle dynamics model and environmental dynamics model are trained separately using different loss functions. The vehicle dynamics model is optimized to accurately predict the ego state based on the prior ego state and the action taken. For this, we employ the Mean Squared Error (MSE) loss, expressed as
\begin{equation}
	\mathcal{L}_{\text{veh}}(\theta) \dot{=} \sum_{t=1}^T \frac{1}{2} \left( f_\theta (s^i_{t-1}, a_t) -s^i_t \right)^2.
\end{equation}
The training process for the environmental dynamics model  closely follows that of the standard RSSM \cite{hafner2023dreamerv3}. Given a batch sequence of observations $o_{1:T}$, ego states $s^i_{1:T}$, rewards $r_{1:T}$, and continuation flags $c_{1:T}$, the parameters of the EDM are optimized end-to-end to minimize the prediction loss $\mathcal L_{\text{pred}}$, prior loss $\mathcal L_{\text{pri}}$, and representation loss $\mathcal L_{\text{rep}}$:
\begin{equation}
	\mathcal L(\phi) = \mathbb E_{q_\phi}\left[ \sum_{t=1}^{T} (\mathcal L_{\text{pred}}(\phi) + \beta_{\text{pri}} \mathcal L_{\text{pri}}(\phi) + \beta_{\text{rep}} \mathcal L_{\text{rep}}(\phi)) \right]
\end{equation}
The prediction loss is used to train the observation decoder, reward predictor, and continuation predictor, and can be expressed as:
\begin{equation}
	\mathcal L_{\text{pred}}(\phi) \dot{=} -\ln p_{\phi}(o_t|z_t,h_t) - \ln p_{\phi}(r_t|z_t,h_t) - \ln p_{\phi}(c_{t}|z_t,h_t).
\end{equation} 
We apply the \textit{symlog} transformation to the decoder and reward predictor, following \cite{hafner2023dreamerv3}, which has proven effective for cross-domain learning. 
The prior loss is used to train a sequential model to predict the next representation by minimizing the Kullback-Leibler (KL) divergence between the predicted and actual next representations. 
The representation loss, on the other hand, encourages representations to become more predictable when the dynamics cannot accurately predict their distribution. These two losses differ in the use of the stop-gradient operator $\text{sg}(\cdot)$ and the loss scale $\beta_{\text{pri}}$ and $\beta_{\text{rep}}$:
\begin{equation}
    \begin{aligned}
        \mathcal L_{\text{pri}}(\phi) &\dot{=} \max( 1,\text{KL}[\text{sg}(q_\phi(z_t|h_t,o_t))&||&p_\phi(z_t|h_t)])  \\
        \mathcal L_{\text{rep}}(\phi) &\dot{=} \max(1,\text{KL}[q_\phi(z_t|h_t, o_t)&||&\text{sg}(p_\phi(z_t|h_t))]).  \\
    \end{aligned}
    \label{eq:kl_balancing}
\end{equation}
We apply free bits \cite{kingma2016improved} by clipping values below 1 to prevent degenerate solutions.

\subsection{Behavior Learning}

\subsubsection{Behavior Model Structure}
Rather than directly learning actions for the ego vehicle, VDD learns goal states that the ego vehicle should reach, and then generates actions through a predefined controller to track these target states. 
This approach offers the advantage of considering long-term objectives and can be integrated with other modules to reduce collision risks \cite{wu2022trajectory}.
Additionally, adjusting the controller enhances the ego vehicle’s robustness to varying vehicle dynamics. 
We employ actor and critic neural networks to train a goal generator, while a PID controller generates the corresponding actions. 
The actor and critic networks derive behaviors solely from abstract sequences predicted by the learned world model.
The actor network takes the environmental state $s^e_t\dot{=}\{h_t,z_t\}$ as the input and outputs a goal state $g_t$. It aims to maximize the expected discounted return $R_t\dot{=}\sum_{\tau=0}^{\infty} \gamma^\tau r_{t+\tau}$ for each model state. The controller receives the goal and the ego vehicle's state $s^i_t$ as inputs and generates an action $a_t$. This feedback control mechanism enhances the robustness of the ego vehicle's behavior against changes in vehicle dynamics. Meanwhile, the critic network learns to predict the return for each state under the current actor and controller:
\begin{equation}
	\begin{aligned}
		&\text{Actor:} &g_t&\sim \pi_\epsilon(g_t|s^e_t) \\
		&\text{Controller:} &a_t&= u_{\kappa}(g_t, s^i_t) \\
		&\text{Critic:}        &v_\xi& \approx \mathbb{E}_{p_\phi,f_\theta, \pi_\epsilon, u_{\kappa}}[R_t],
	\end{aligned}
\end{equation}
where $\pi, \kappa, \xi$ are the parameters of the components. During the training process, the agent starts from the latent representations of replayed inputs and produces imagined trajectories using the dynamics predictor, actor network and controller. We compute bootstrapped $\lambda\text{-returns}$ that integrate the predicted rewards and values:
\begin{equation}
    \begin{aligned}
        R^\lambda_t  &\dot{=} r_t + \gamma c_t \left( (1-\lambda)v_\xi(s_{t+1}) + \lambda R^\lambda_{t+1} \right) \\
        R^\lambda_{T} & \dot{=} v_\xi(s_{T})
    \end{aligned}
\end{equation}

\subsubsection{Critic Learning} 
The learning objective of the critic is to minimize the discrepancy between the predicted values of model states and their expected cumulative returns. Following DreamerV3, we employ a discrete regression method to train the critic using two-hot encoded target values.
Specifically, we first transform the $\lambda\text{-returns}$ using the symlog function and then discretize the output range into a sequence $B$ of $K$ equally spaced bins, denoted as $b_i$. These values are subsequently two-hot encoded to form a soft label for the softmax distribution predicted by the critic network, represented as
\begin{equation}
	y_t = \text{sg}(\text{twohot}(\text{symlog}(R^\lambda_t))),
\end{equation}
where $\text{sg}(\cdot)$ is the stop-gradient operator, $\text{symlog}(\cdot)$ is the symlog transformation, and $\text{twohot}(\cdot)$ is the two-hot encoding function. 
The critic network is trained by minimizing the categorical cross-entropy loss between the predicted soft labels and the true labels:
\begin{equation}
	\mathcal L_{critic}(\xi) \dot{=} -\sum_{t=1}^T y_t^\top \ln p_\xi(\cdot|s_t).
\end{equation}
The critic network can predict any continuous value within range of $B$ using the symexp transform:
\begin{equation}
	v_\xi(s_t) \dot{=} \text{symexp}(p_\xi(\cdot|s_t)^\top B).
\end{equation}

\subsubsection{Actor Learning}
The actor network is optimized to produce goals that maximize expected returns while balancing exploration, supported by an entropy regularizer. These goals are also designed to be attainable within a set time budget using the predefined controller. To train the actor network, we backpropagate value gradients through sequences of sampled model states and actions. To ensure stable return scaling, we adopt a method that reduces large returns without amplifying smaller ones, following the approach of DreamerV3 \cite{hafner2023dreamerv3}. Additionally, a penalty term is incorporated to encourage the actor to generate goals reachable within the specified time limit. Consequently, the actor network's loss function is formulated as: 
\begin{equation}
	\begin{aligned}
		\mathcal L_{\text{actor}}(\xi)\dot{=} & \sum_{t=1}^{T}\mathbb{E}_{p_\phi,f_\theta, \pi_\epsilon, u_{\kappa}}\left[ \text{sg}(R^\lambda_t)/\text{max}(1,S) \right] \\ 
		& - \omega_{\text{ent}} \mathrm{H} \left[ \pi_\epsilon(g_t|s^e_t) \right] + \omega_{\text{reach}}(g_t-\hat{s}^i_{t+h})^2,
	\end{aligned}
\end{equation}
where $\omega_{\text{ent}}$ and $\omega_{\text{reach}}$ are the entropy and reachability penalty scales, and $\mathrm{H}(X) = \mathbb{E}[-\ln(p(X))]$ represents the entropy. To normalize returns while being robust to outliers in the randomized environment, we use an exponentially decaying average of the range from the 5th to 95th batch percentile of the returns, i.e. $S=\text{Per}(R^\lambda_t, 95) - \text{Per}(R^\lambda_t, 5)$.
  
\subsection{Policy Adjustment during Deployment}
\begin{figure}[htbp]
	\centering
	\includegraphics[width=0.45\textwidth]{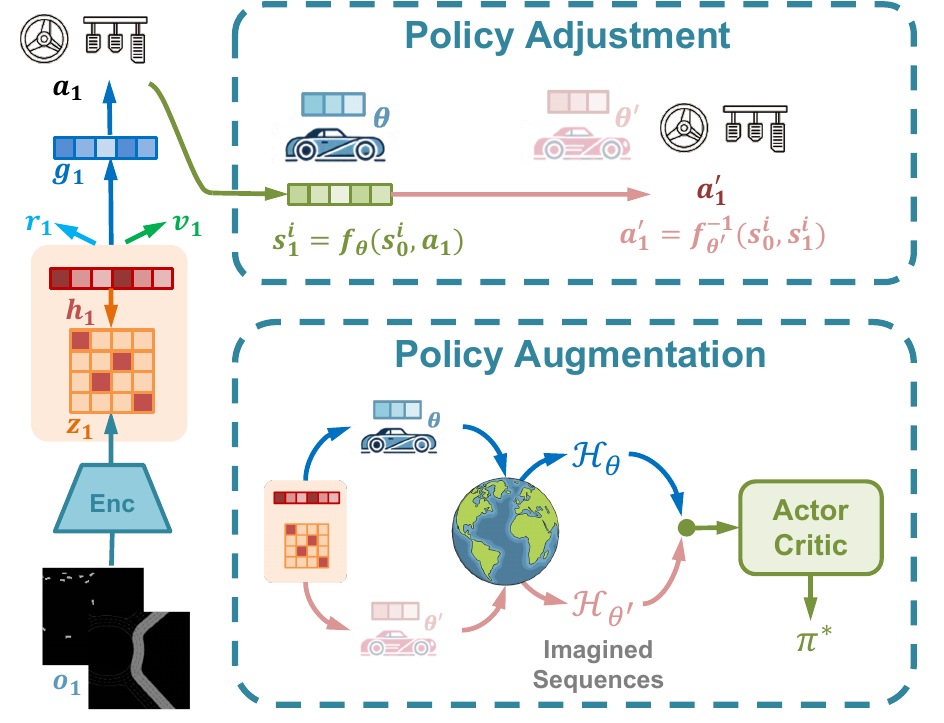}
	\caption{Policy adjustment (top) and policy augmentation (bottom) strategies to improve the robustness to vehicle dynamics shifts.}
	\label{fig:policy_adapt_and_augment}
\end{figure}

The core idea of Policy Adjustment during Deployment (PAD) is to directly adjust the policy learned in the original environment when deploying it in a shifted environment. The objective of this adjustment is to enable the vehicle to achieve the same target states in the shifted environment as it would in the original environment.
As shown at the top of Fig.~\ref{fig:policy_adapt_and_augment}, when deploying on a vehicle with new dynamics parameters $\theta^{'}$, we begin by predicting the vehicle’s state using the original policy and vehicle dynamics model. This predicted state, combined with inverse dynamics, is then used to generate the adjusted policy, represented as:
\begin{equation}
	a^{'}_t = f^{-1}_{\theta^{'}}(s^i_{t-1},f_\theta(s^{i}_{t-1},a_t))
\end{equation}
PAD provides a straightforward and effective approach to managing shifts in vehicle dynamics by ensuring that the policy achieves consistent states across both the original and shifted environments, thereby maintaining the same return as under the original conditions. 
However, when a vehicle with modified dynamic properties cannot achieve the same states as the original vehicle, relying solely on policy adjustment may not yield equivalent returns to those under the original dynamics. For instance, if a vehicle is deployed with a maximum front-wheel steering angle smaller than that of the training vehicle, policy adjustment alone cannot map its states to align with those attainable under the original dynamics.

\subsection{Policy Augmentation during Training}
Policy Augmentation during Training (PAT) focuses on enhancing the training data by using world models to simulate trajectories across a range of vehicle dynamics parameters. This technique supports the optimization of policies to better adapt to shifts in vehicle dynamics.
As is shown at the bottom of Fig.~\ref{fig:policy_adapt_and_augment}, when the agent generates imagined trajectories from representations of replayed inputs, it can utilize both the true vehicle parameter $\theta$ and the shifted parameter $\theta^{'}$ to create distinct trajectories, represented as $\mathcal H_\theta$ and $\mathcal H_{\theta^{'}}$, respectively. 
We then combine these trajectories to train the actor and critic networks. Specifically, we treat the vehicle dynamics parameters as contextual information and augment the world model state to create a context-aware state, defined as $s^c_t = \left\{s^e_t, ctx_{\theta^{'}}\right\}$. Here, the context information $ctx_{\theta^{'}}$ is defined as the ratio of the shifted parameters to the original parameters, i.e., $ctx_{\theta^{'}} = \frac{\theta^{'}}{\theta}$. For the original vehicle parameters, this context vector consists of all elements equal to one, represented as $ctx_{\theta} = [1,1,...,1]^\top \in \mathbb{R}^n$, where $n$ denotes the dimension of the vehicle parameters. In this framework, the critic network is trained to predict the value of the model state under both the original and shifted vehicle dynamics parameters:
\begin{equation}
	\begin{aligned}
		\mathcal L_{\text{critic}}(\xi) \dot{=} -\sum_{t=1}^{T} & [(1-\omega_{\text{cont}})y_{t,ctx_\theta}^\top\ln p_\xi(\cdot|s^e_t, ctx_\theta) + \\ 
		& \omega_{\text{cont}}y_{t,ctx_{\theta^{'}}}^\top\ln p_\xi(\cdot|s^e_t, ctx_{\theta^{'}})   ],
	\end{aligned}
\end{equation}
where $\omega_{\text{cont}}$ is the weight of the critic loss calculated by the shifted context information. Then the actor network $\pi_{\epsilon}$ is trained to maximize the expected returns under both the original and shifted vehicle dynamics parameters.

\section{Experiments}
\label{experiments}
In this section, we evaluate the proposed method's performance and robustness in response to changes in ego vehicle dynamics.

\subsection{Experiment Setup}
\subsubsection{Simulation environment}
\begin{figure}[h]
    \centering
	\includegraphics[width=0.4\textwidth]{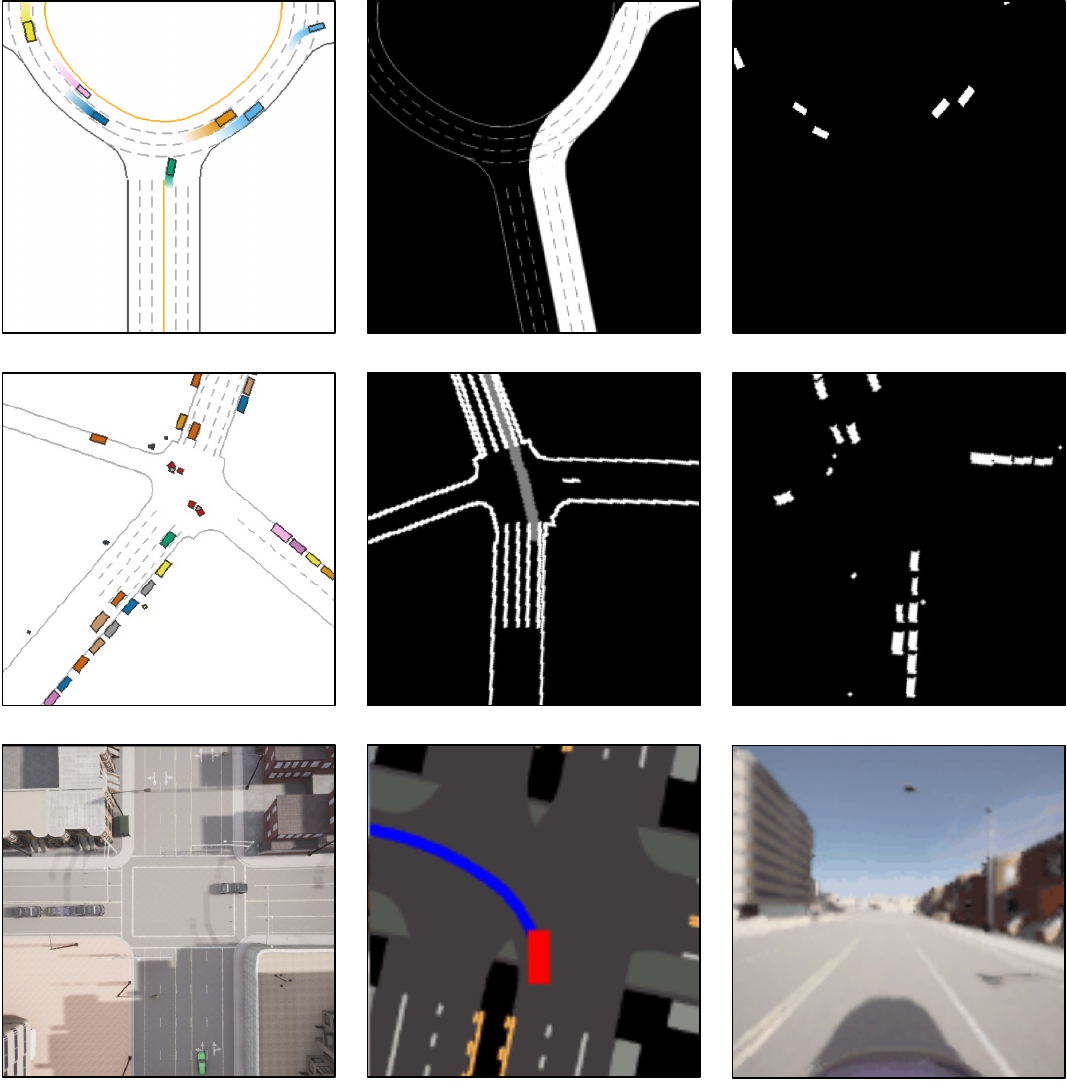}
	\caption{ {Maps visualization (left column) and observations (central and right columns) used in the experiments. 
			The first two rows show the visualization of roundabout (top row) and example from the nuPlan dataset (middle row) in the MetaDrive simulator.
			Each BEV observation frame consists of two channels: one representing the road geometry and navigation information (central column) and the other representing surrounding vehicles (right column). The bottom row shows the visualization of the intersection in the CARLA simulator. Each observation frame consists of a route map image and a front camera image.}}
			
	\label{fig:sensors_input}
\end{figure}

To assess the performance of our method, we use MetaDrive \cite{li2022metadrive}, a highly compositional autonomous driving simulator, to conduct extensive experiments. 
MetaDrive provides a wide range of driving scenarios through procedural generation and supports varying ego vehicle dynamics settings. We select two road maps, roundabout and intersection, to test the effectiveness of our approach, setting the traffic density at 0.35 to simulate a dense urban traffic environment.
In each scenario, the ego vehicle starts in a random lane with a randomly assigned destination. Traffic vehicles are randomly generated in the environment at a predefined density and are managed by MetaDrive’s rule-based traffic manager.   {To further validate the effectiveness of the proposed model on real-world traffic data, we convert the Pittsburgh's traffic data in the nuPlan \cite{nuplan} dataset using ScenarioNet \cite{li2023scenarionet} and import it into the MetaDrive simulator.  We use 8000 scenario logs for training and test the model on additional 100 random scenarios.}
The ego vehicle’s objective is to navigate through traffic and reach the destination safely within a given time budget, avoiding collisions and staying on the roadway. 
The ego vehicle’s sensory input consists of a 2-channel bird's-eye-view (BEV) image and kinematic data. The BEV image, as shown in Fig.~\ref{fig:sensors_input}, includes two channels: one for road geometry and one for surrounding vehicles, each with dimensions of $128 \times 128 \times 1$. The BEV image can be obtained through various perception techniques \cite{li2022bevformer}. As this work focuses on decision-making, we assume the BEV image is available as a sensory input. The kinematics information includes the ego vehicle's pose, velocity, yaw rate and acceleration, denoted as $[x,y,\theta, v, \omega, a]$, which are readily obtainable from onboard sensors.

 {Moreover, although end-to-end autonomous-driving policies that take BEV as input have already gained widespread acceptance \cite{chen2019modefreerl}, we conduct additional training and evaluation on the CARLA \cite{carla} simulator to demonstrate the effectiveness of our method when raw camera data are used. We train the VDD model on the unprotected left-turn task from CarDreamer  \cite{cardreamer}. In this setup, the ego vehicle’s observations consist of a $128\times128\times3$ RGB route map and front-view camera images. We modify the VDD encoder and decoder: the encoder receives the two observations stacked along the channel dimension, while the decoder reconstructs both the front-camera view and a BEV image that includes the surrounding vehicle’s position.}

The reward function of the MetaDrive environment is composed of three parts as follows:
\begin{equation}
	R = c_1 R_{prog} + c_2 R_{speed} + R_{term}.
\end{equation}
The \textit{progress reward} $R_{prog} = d_t - d_{t-1}$, where $d_t$ represents the ego vehicle’s longitudinal movement in Frenet coordinates towards the destination, encourages forward movement.
The \textit{speed reward} $R_{speed} = v / v_{max}$ incentivizes the agent to maintain a high speed, where $v$ and $v_{max}$ denote the current and maximum speeds, respectively.
Additionally, we define a sparse \textit{terminal reward}, which is non-zero only upon episode termination. For the experiments, we set $c_1 = 1$, $c_2 = 0.1$, and $v_{max} = 50 \text{km/h}$. The agent receives a reward of 10 upon reaching the destination, a penalty of -10 for going off-road, and -20 for collisions. The agent does not have direct access to the reward function. It only receives the reward value provided by the environment for each observation.

\subsubsection{Baselines and evaluation metrics}
We compare the performance of our method with the following baseline methods:
\begin{itemize}
	\item {\bf{IDM}}\cite{treiber2000congested}: An agent based on Intelligent Driver Model (IDM), a rule-based car-following model.
	\item {\bf{SAC}} \cite{haarnoja2018soft}: An agent trained from scratch using Soft Actor-Critic (SAC), a typical model-free off-policy deep RL algorithm.
	\item {\bf{PPO}} \cite{schulman2017proximal}: An agent trained from scratch using Proximal Policy Optimization (PPO), a model-free on-policy DRL algorithm.
	\item {\bf{DreamerV3}}\cite{hafner2023dreamerv3}: An agent trained through DreamerV3, a model-based RL method.
\end{itemize}

To reflect the performance of AVs from different aspects, we use the following evaluation metrics \cite{wang2023asaprl}:
\begin{itemize}
	\item {\bf{Episode reward}}: the cumulative sum of rewards obtained within a single episode.
	\item {\bf{Success rate}}: the percentage of episodes in which the agent reaches the destination within the allotted time.
	\item {\bf{Collision rate}}: the percentage of episodes that involve a collision.
	\item {\bf{Route completion ratio}}: the proportion of the route completed by the ego vehicle relative to the total route length.
\end{itemize}

\subsubsection{Implementation details}
For the environment dynamics model, we employ a 5-layer CNN encoder with filter sizes of (32, 64, 128, 256, 512) and a kernel size of 4 and stride of 2 in each layer. Each convolutional layer is followed by a normalization layer and a SiLU activation function. The stochastic state is modeled by a categorical distribution with 32 classes, while a GRU with 512 hidden units represents the deterministic state. The image decoder consists of 5 transposed convolutional layers with filters (256, 128, 64, 32, 2). The reward predictor, continuation predictor, actor, and critic networks are MLPs with 512 hidden units and a SiLU activation function, except for the output layer. We select the acceleration and yaw rate of the ego vehicle as inputs for the environment dynamics model.
We collect data by running the simulator with various random seeds, storing it in a replay buffer with a capacity of $2 \times 10^5$. During each training step, we sample a mini-batch of size 16 with a sequence length of 64 to update the model parameters. The model is trained using the Adam optimizer, with the environment world model’s learning rate set to $10^{-4}$ and a gradient clipping threshold of 1000. The learning rate for the actor-critic is set to $3 \times 10^{-5}$, with a gradient clipping threshold of 100.  {We use identical values for the above parameters in both the MetaDrive and CARLA experiments.}.

\subsubsection{Experiment settings}
To evaluate the performance of VDD, we save the model every $10^5$ environment steps and test it with random seeds that differ from those used during training. 
For each test, we use three sets of identical random seeds, with each set consisting of 100 episodes. We record both the mean and variance of the evaluation metrics.
To further evaluate the robustness of the method to varying vehicle dynamics parameters, we test the trained models with different sets of vehicle dynamics parameters. Specifically, we adjust the mass and maximum steering angle of the ego vehicle by applying multiplicative scaling factors in the MetaDrive simulator. We consider a grid of values in ${0.5, 0.75, 1.0, 1.25, 1.5}$ to scale the vehicle parameters relative to those in the original training environment (mass = 1100 kg, max steering angle = 40 degrees). Each parameter set is tested over 100 episodes. 

\subsection{Results and Discussions}
In this subsection, we first highlight the advantages of the proposed method from a macro perspective using the previously discussed evaluation metrics. We then provide several micro-level indicators and case studies to illustrate the method’s functionality in greater detail.

\subsubsection{Driving Performance}
\begin{figure*}[h]
    \begin{minipage}[b]{0.8\textwidth}
        \centering
        \includegraphics[width=7.5in]{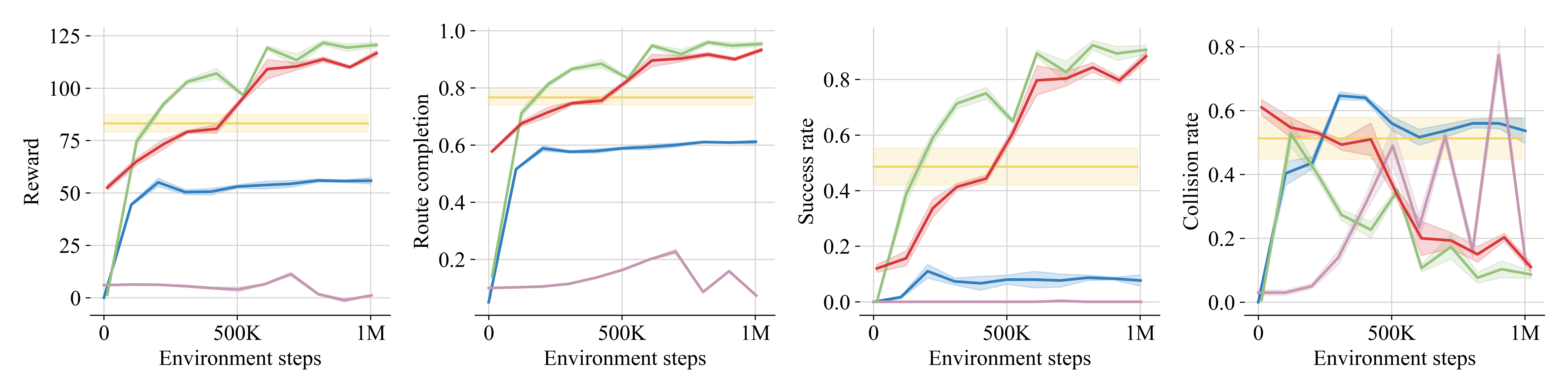}
    \end{minipage}
    \hspace{0.05\textwidth} 
    \begin{minipage}[b]{0.8\textwidth}
        \centering
        \includegraphics[width=7.5in]{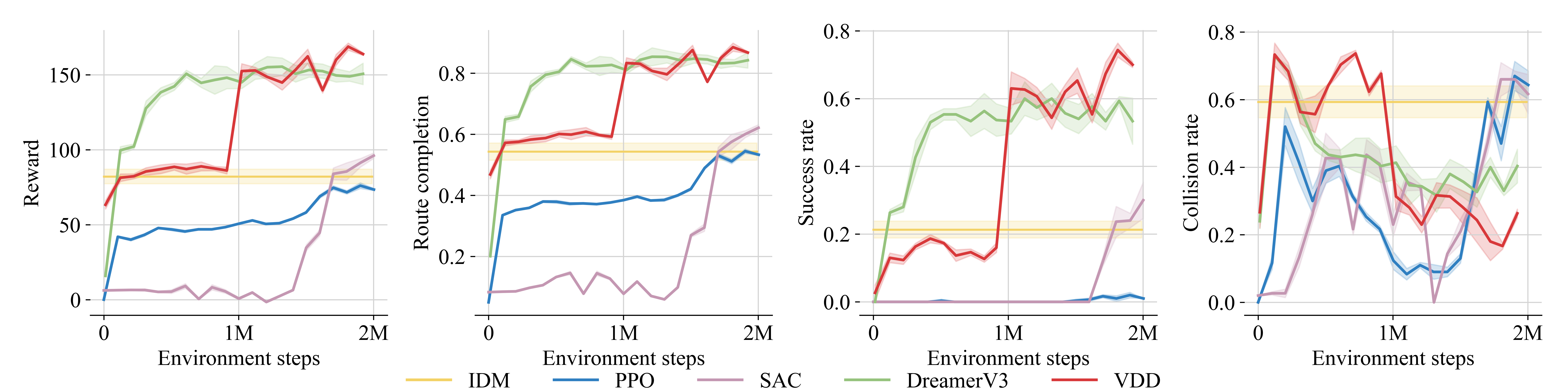}
    \end{minipage}
	\caption{Performance comparison of various methods on the intersection (top) and roundabout (bottom) maps. The curves depict evaluation results for checkpoints over 100 episodes at intervals of every $10^5$ environment steps. Solid lines represent the mean results across three trials, while shaded areas denote the standard deviation.}
	\label{fig:performance_comparison}
\end{figure*}

We begin by evaluating and comparing the driving performance of various methods on the intersection and roundabout maps. The curves in Fig.~\ref{fig:performance_comparison} illustrate the change in model performance with respect to the environment steps, and Tab.~\ref{tab:results} presents the final performance of each method upon completion of training. 
Although the IDM method has demonstrated high scores in closed-loop testing \cite{dauner2023parting}, it performs poorly in dense and uncertain traffic environments. Model-free RL methods, such as PPO and SAC, exhibit a relatively slow increase in cumulative rewards as data collection progresses, failing to converge to high performance levels within the allocated training steps. This aligns with findings reported in previous studies \cite{zhou2023accelerating, wang2023asaprl}.
In contrast, the model-based RL method DreamerV3 quickly converges to a high total reward, showcasing the high sample efficiency characteristic of world-model-based learning approaches. Comparatively, the proposed VDD method converges at a slightly slower rate on the intersection map. On the roundabout map, it initially remains at a locally optimal policy for the first one million (1M) environment steps but ultimately achieves higher cumulative rewards and success rates by 2M steps.
We attribute the slower convergence rate of the VDD method to the increased complexity of goal learning compared to direct action learning. Nonetheless, the integration of learned goals with controller execution in VDD enhances overall driving performance.
As shown in Tab.~\ref{tab:results}, although VDD’s performance is slightly lower than DreamerV3’s on intersection maps, it demonstrates a notable advantage on the more challenging roundabout maps, achieving a 17\% improvement in success rate.
Additionally, as illustrated in Fig.~\ref{fig:action_comparison}, this approach results in smoother action transitions than models that learn actions directly, thereby enhancing driving comfort.  {In addition, we conduct supplementary experiments to verify that more accurate vehicle-dynamics parameters, obtained through online learning, help maintain the robustness of model performance. The details and results of the experiment are presented in the \hyperref[appendix]{Appendix}.}

 {We also report the experimental results on the nuPlan dataset and the CARLA simulator in Tab.~\ref{tab:results}. It can be seen that, even when raw camera images are used as input, VDD still outperforms DreamerV3. However, results on nuPlan reveal that VDD performs slightly worse than DreamerV3, and both lag behind the rule-based IDM controller. We attribute this to the fact that the IDM controller uses precise vector information as input, thereby bypassing image processing. We will make specific effort to improve VDD's performance on real-world datasets in the future work.}

\begin{table*}
    \centering
    \caption{ {Performance comparison of different methods}}
    \begin{tabular}{p{2cm}cccccccccccccc}
        \hline
        \multirow{3}{*}{Methods} 
        & \multicolumn{11}{c}{MetaDrive Simulator} & \multicolumn{3}{c}{CARLA Simulator}\\
        & \multicolumn{4}{c}{T-Intersection} & \multicolumn{4}{c}{Roundabout}& \multicolumn{3}{c}{nuPlan} & \multicolumn{3}{c}{Intersection}\\
        \cline{2-13}
        & RW$\uparrow$ & RC$\uparrow$ & SR$\uparrow$ & CR $\downarrow$ & RW$\uparrow$ & RC$\uparrow$ & SR $\uparrow$ & CR $\downarrow$& RW$\uparrow$ &  SR $\uparrow$ & CR $\downarrow$ & RW$\uparrow$ &  SR $\uparrow$ & CR $\downarrow$\\
		\hline
		
        IDM & 83.22 & 0.77 & 0.49 & 0.51 
			& 82.17 & 0.54 & 0.21 & 0.59 &\textbf{98.62} &\textbf{0.93} &\textbf{0.03} & 212.25 & 0.29 & 0.71\\

        SAC & 1.11 & 0.07 & 0.00 & 0.12 
			& 96.08 & 0.62 & 0.30 & 0.62 \\
  
        PPO & 55.83 & 0.61 & 0.08 & 0.54 
			& 73.44 & 0.53 & 0.01 & 0.64 \\
   
        DreamerV3 & \textbf{120.63} & \textbf{0.95} & \textbf{0.91} & \textbf{0.09}
			      & 150.64 & 0.84 & 0.53 & 0.40 &78.38 &0.79 &0.17 & 423.06 & 0.91 & 0.09\\
    
        \textbf{VDD(ours)} & 116.69 & 0.93 & 0.88 & 0.11 
		    & \textbf{163.82} & \textbf{0.87} & \textbf{0.70} & \textbf{0.26} &72.29 &0.77 &0.19 & \textbf{427.18} & \textbf{0.94} & \textbf{0.06} \\
        \hline
        \par \small
      \mbox{\footnotesize \textbf{*} RW: Reward, RC: Routine completion, SR: Success rate, CR: Collision rate}
    \end{tabular}
	\label{tab:results}
\end{table*}

\subsubsection{Ablation Analysis}
\begin{figure}
	\includegraphics[width=0.48\textwidth]{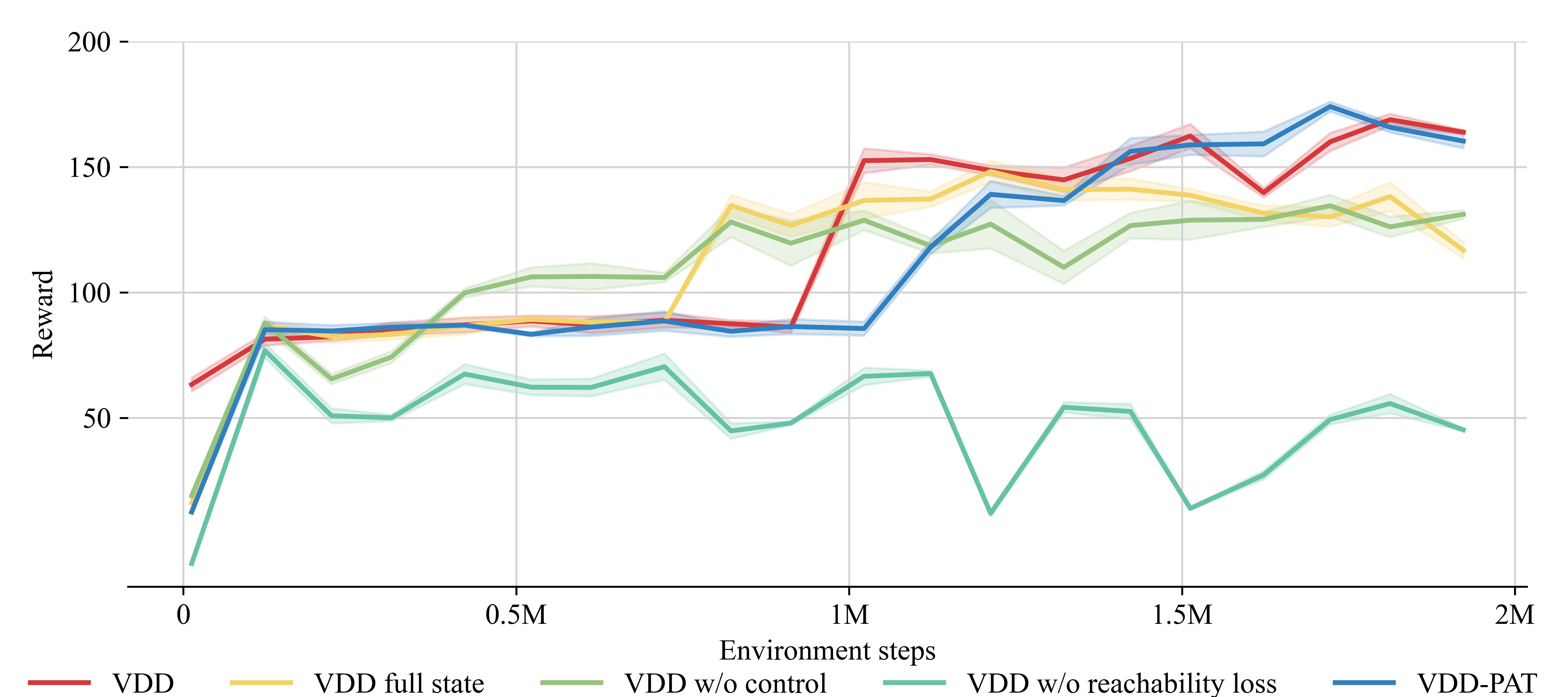}
	\caption{Ablation analysis of different designs  in VDD in the roundabout map.}
	\label{fig:ablation}
\end{figure}

In this subsection, we present the results of the ablation study, examining the impact of key components in the VDD method, as illustrated in Fig.~\ref{fig:ablation}.
\begin{itemize}
	\item Influence of environment dynamics input: We compare the performance of the model using all ego vehicle states as inputs versus VDD that uses only acceleration and yaw rate. The results indicate that VDD achieves better performance.
	\item Effect of explicit controller modeling: In VDD, we integrate the ego vehicle’s tracking controller for generating actions and imagined trajectories. To evaluate the impact of this design choice, we test a variant without controller modeling. The absence of controller modeling leads to a slight decrease in performance, suggesting its beneficial role in the model.
	\item Impact of reachability loss: We further explore the role of reachability loss by comparing models trained with and without it. The results reveal that omitting reachability loss causes a significant drop in performance, underscoring its importance in guiding the model to learn reachable goals.
	\item Influence of using PAT: Finally, we assess the effect of incorporating the PAT strategy on VDD’s performance. Results indicate minimal impact on performance in environments with the original vehicle dynamics parameters. However, due to simultaneous policy learning across varied hypothetical vehicle dynamics, the model’s convergence rate to the optimal policy is slightly reduced.
\end{itemize} 

\subsubsection{ {Sensitivity Analysis}}
\begin{figure}
	\includegraphics[width=0.5\textwidth]{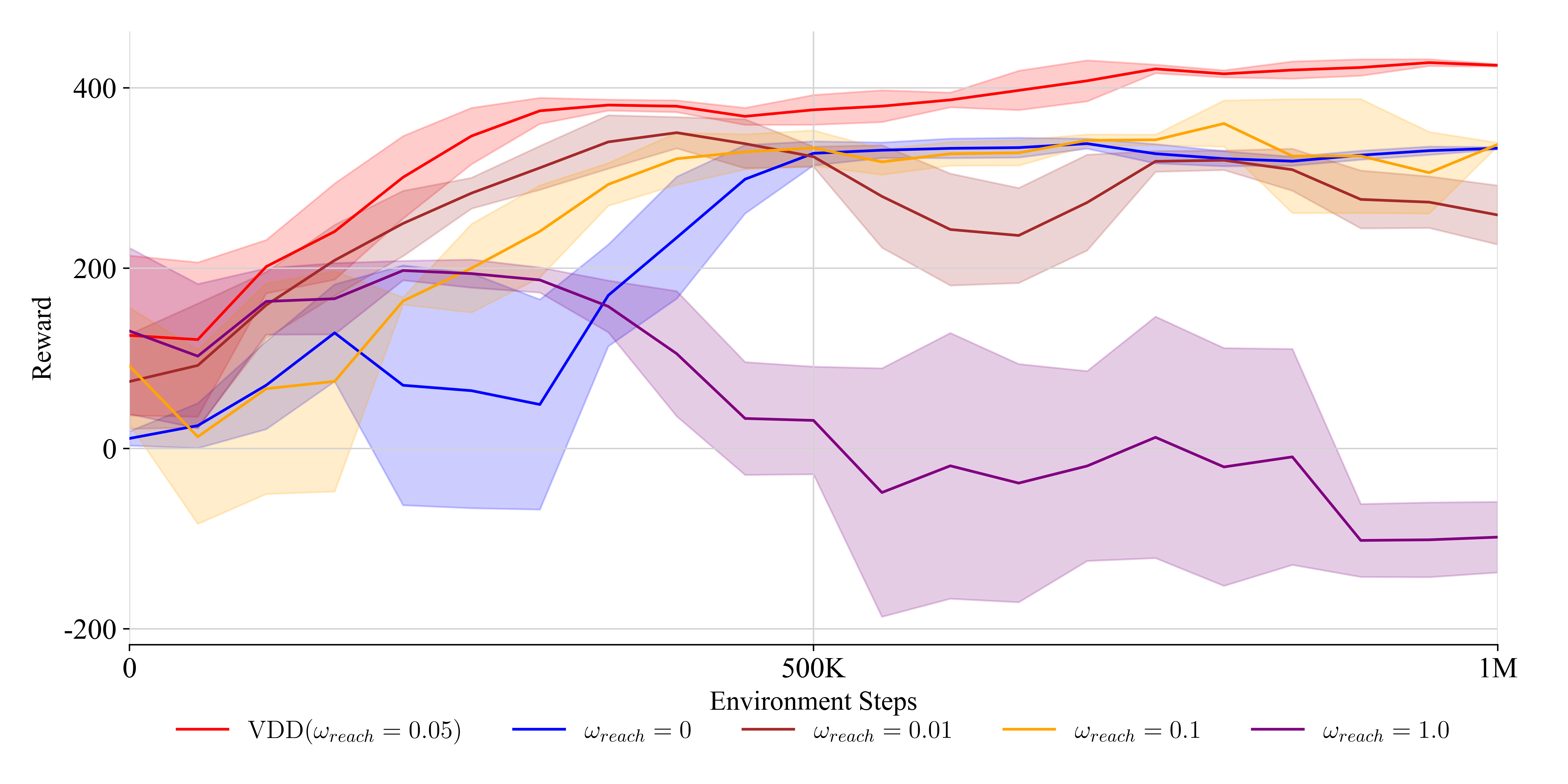}
	\caption{ {Performance of VDD of the unprotected left turn task in the CARLA simulator and sensitivity analysis of the reachability penalty $\omega_{\text{reach}}$.}}
	\label{fig:sensitivity}
\end{figure}

 {
In this subsection, we analyze the sensitivity of the proposed model to its hyper-parameters. We divide the parameters in VDD into two groups: those shared with the baseline DreamerV3, and those introduced by VDD’s novel design.
DreamerV3 is a general algorithm that can solve a wide range of domains with fixed hyper-parameters, so we adopted most of them directly from the original paper \cite{hafner2023dreamerv3}. Additionally, we empirically identify and tune several critical hyper-parameter, such as replay capacity and batch size.  Experimental results show that this choice yields strong performance on autonomous driving tasks. In this study, we keep the hyper-parameters in the VDD model identical to those of baseline, thereby demonstrating the effectiveness of our novel design. During VDD training, the reachability penalty $\omega_\text{reach}$ has a significant impact on model performance. We evaluate different magnitudes of this penalty on the unprotected left-turn task in CARLA, and the results are shown in Fig.~\ref{fig:sensitivity}. An excessively large penalty causes the model to collapse, whereas an overly small one slows convergence. Therefore, when selecting the penalty value, we recommend starting with a relatively small value and increasing it exponentially to identify the correct order of magnitude, then performing finer-grained tuning to achieve optimal performance. }

\subsubsection{Robustness to Vehicle Dynamics Shifts}
 
\begin{figure*}
	\centering
	\begin{minipage}[b]{0.45\textwidth}
		\centering
		\includegraphics[width=\textwidth]{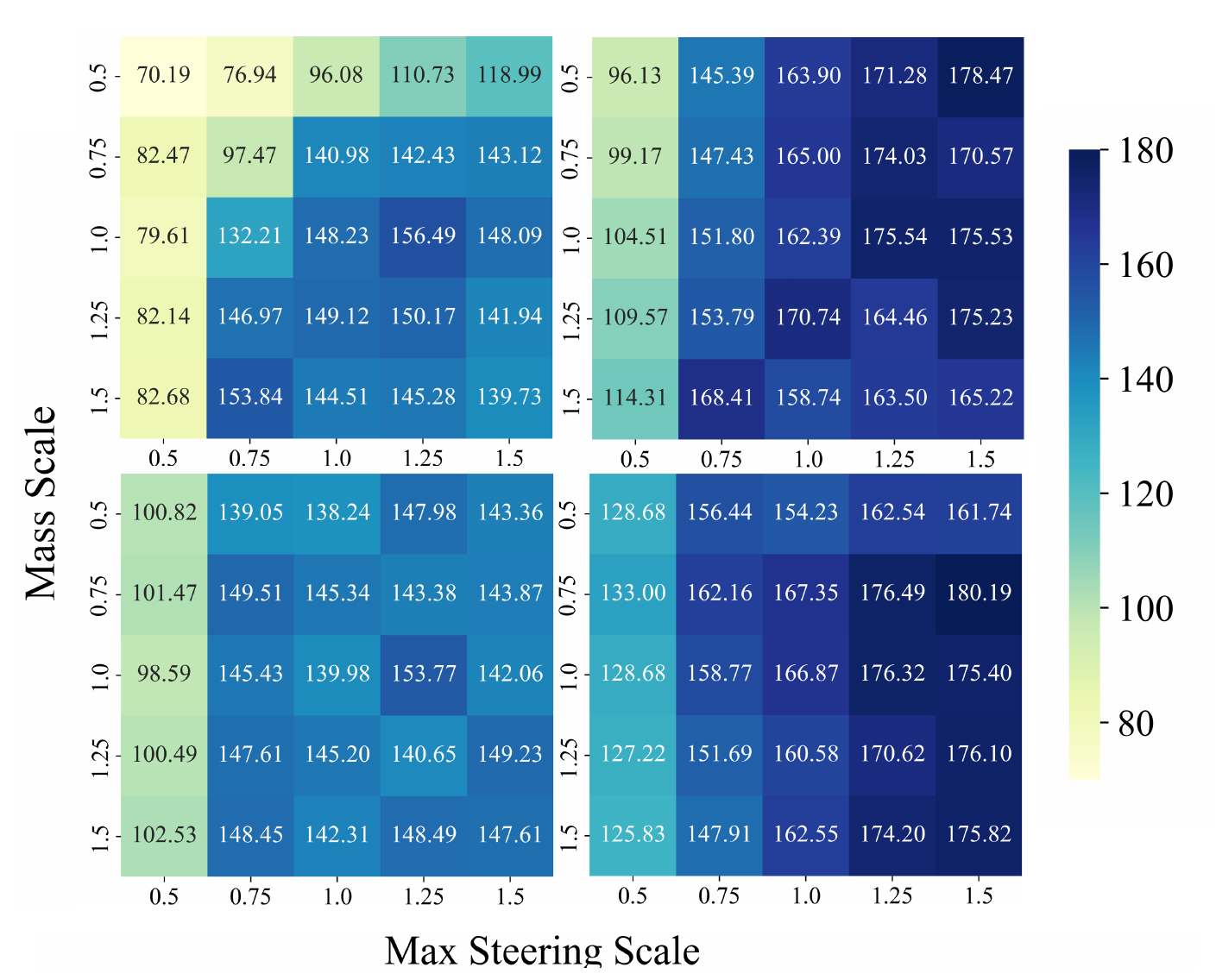}
	\end{minipage}
	\hspace{0.05\textwidth} 
	\begin{minipage}[b]{0.45\textwidth}
		\centering
		\includegraphics[width=\textwidth]{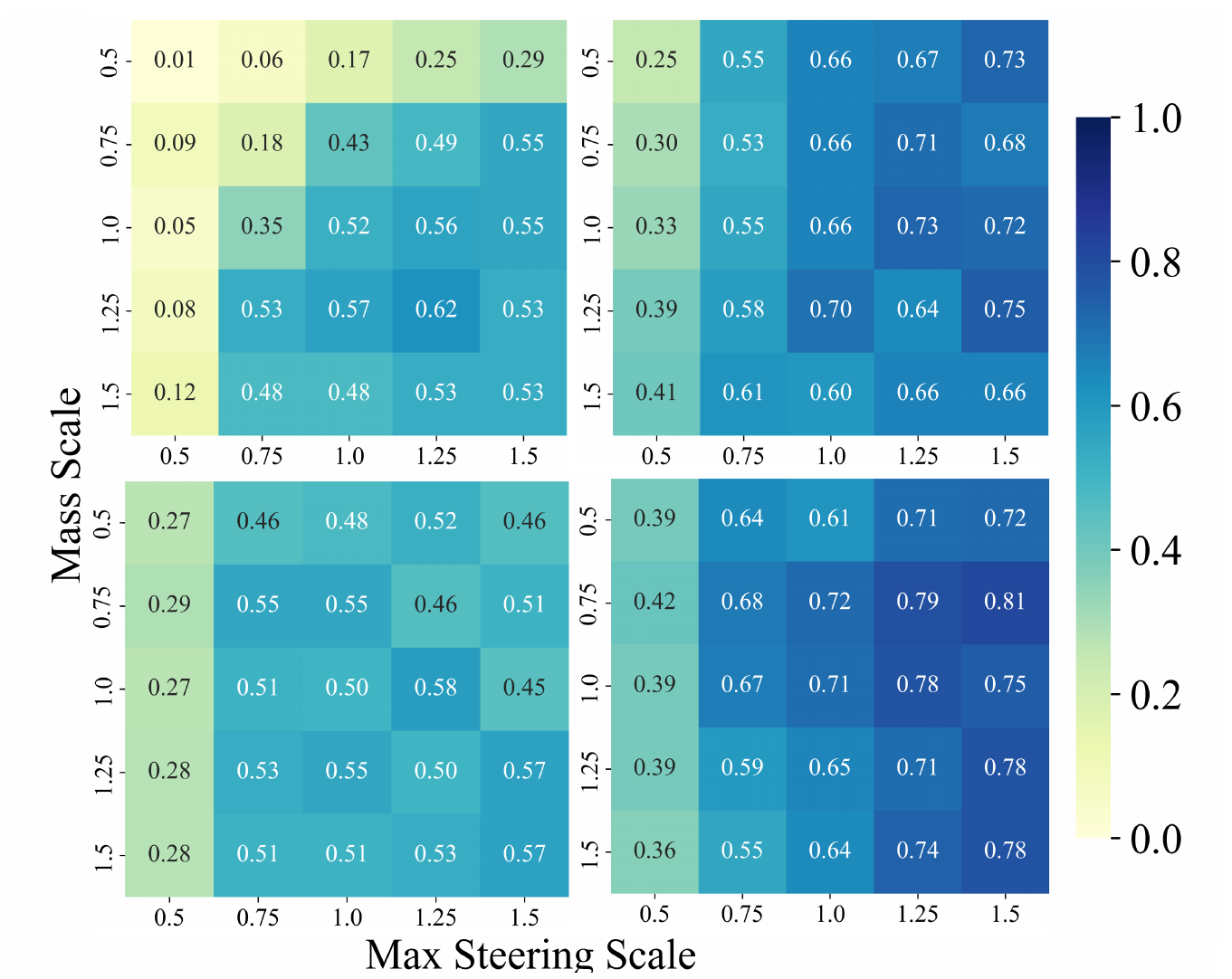}
	\end{minipage}
	\caption{ {Average episodic rewards (left)  and success rate (right) over 100 episodes for DreamerV3 (top left), DreamerV3-PAD (bottom left), VDD (top right) and VDD-PAT (bottom right) under varying vehicle dynamics parameters. The central cell (1,1) represents the in-sample data configuration.}}
	\label{fig:varying_dynamics}
\end{figure*}

{We display the average episodic rewards and success rate of different methods under varying vehicle dynamics shifts in Fig.~\ref{fig:varying_dynamics}. 
Using a $5\times5$ grid, we record the mean performance over 100 episodes for each parameter configuration. 
The x-axis denotes the scaling of the maximum steering angle, while the y-axis represents the mass scaling. 
The central cell (1,1) corresponds to the parameters used in the training environment. }
As can be seen in the figure, it is evident that the DreamerV3 (top left) approach lacks robustness to changes in vehicle dynamics parameters. While performance remains stable for some parameter configurations near the center, it degrades significantly in most dynamics-shifted environments, showing a pronounced sensitivity to both steering angle and mass changes.
In contrast, DreamerV3-PAD (bottom left) can partially mitigate the impact of shifted dynamics by adjusting actions from the trained policy based on the vehicle dynamics model. This adaptation shows greater effectiveness for mass shifts than for maximum steering angle shifts. A reduced maximum steering angle constrains the vehicle’s ability to recover to its desired state using adjustments alone, highlighting that merely adapting the control system to align with policy training under different vehicle dynamics lacks broad generalizability. Adjustments in the planning policy are essential to maintain performance under such dynamics shifts.
Compared to DreamerV3, VDD (top right) demonstrates greater robustness to changes in vehicle dynamics parameters. Specifically, VDD’s performance is stable across mass variations and even improves in configurations with larger maximum steering angles. This robustness can be attributed to two main factors: (1) The use of parameter-independent ego-state inputs and a decoupled world model minimizes the impact of vehicle-specific changes on environmental dynamics, and (2) The actor network’s goal prediction remains independent of vehicle attributes, with the addition of a feedback controller enhancing resilience to variations in vehicle parameters. Nonetheless, similar to DreamerV3, VDD’s performance is affected by the limitations of reduced maximum steering angles. We apply the PAT strategy, using imagined data to train policies under conditions with restricted maximum steering angles. Experimental results indicate that the VDD-PAT (bottom right) method achieves significantly higher cumulative rewards compared to the VDD method when the maximum steering angle scale is set to 0.5, demonstrating the effectiveness of the PAT strategy.
 {The right side of the Fig.~\ref{fig:varying_dynamics} displays the success rate under vehicle dynamics shift. It can be seen that the VDD-PAT approach maintains a higher success rate across different parameter settings. It can also serve as a reference for exploring the boundary of robustness of the proposed method.}

\subsubsection{Case Study}
\begin{figure}
	\centering
	\includegraphics[width=0.45\textwidth]{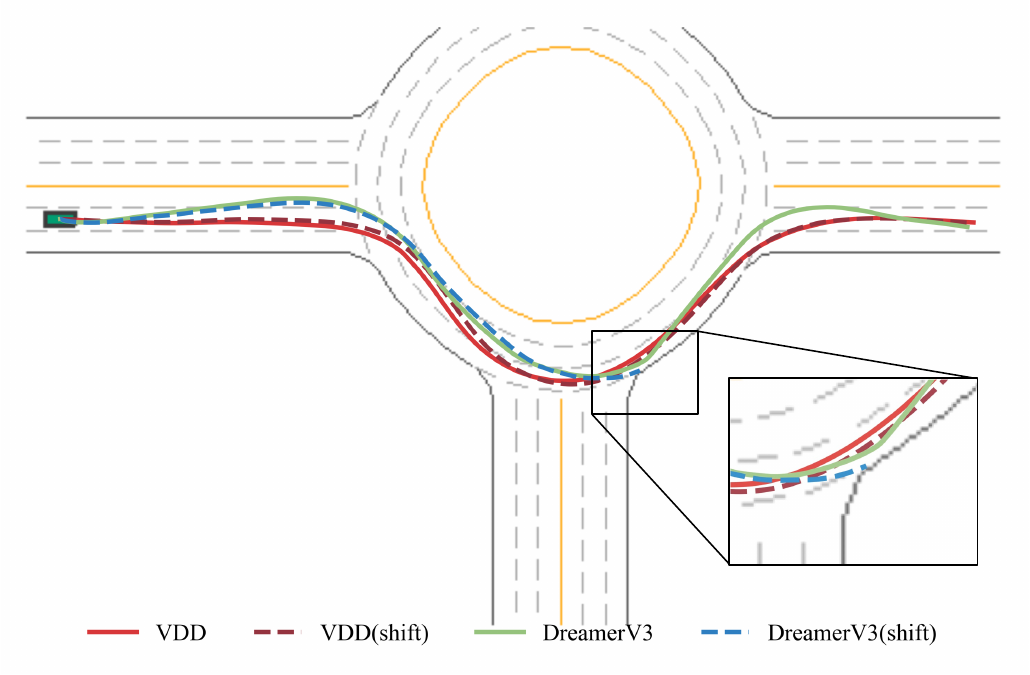}
	\caption{Trajectories for VDD and DreamerV3 during test episodes. Solid lines indicate the trajectories in the original environment, while dashed lines represent the shifted environment. Other vehicles present in the episodes are omitted for clarity.}
	\label{fig:traj_comparison}
\end{figure}

\begin{figure}
	\centering
	\includegraphics[width=0.45\textwidth]{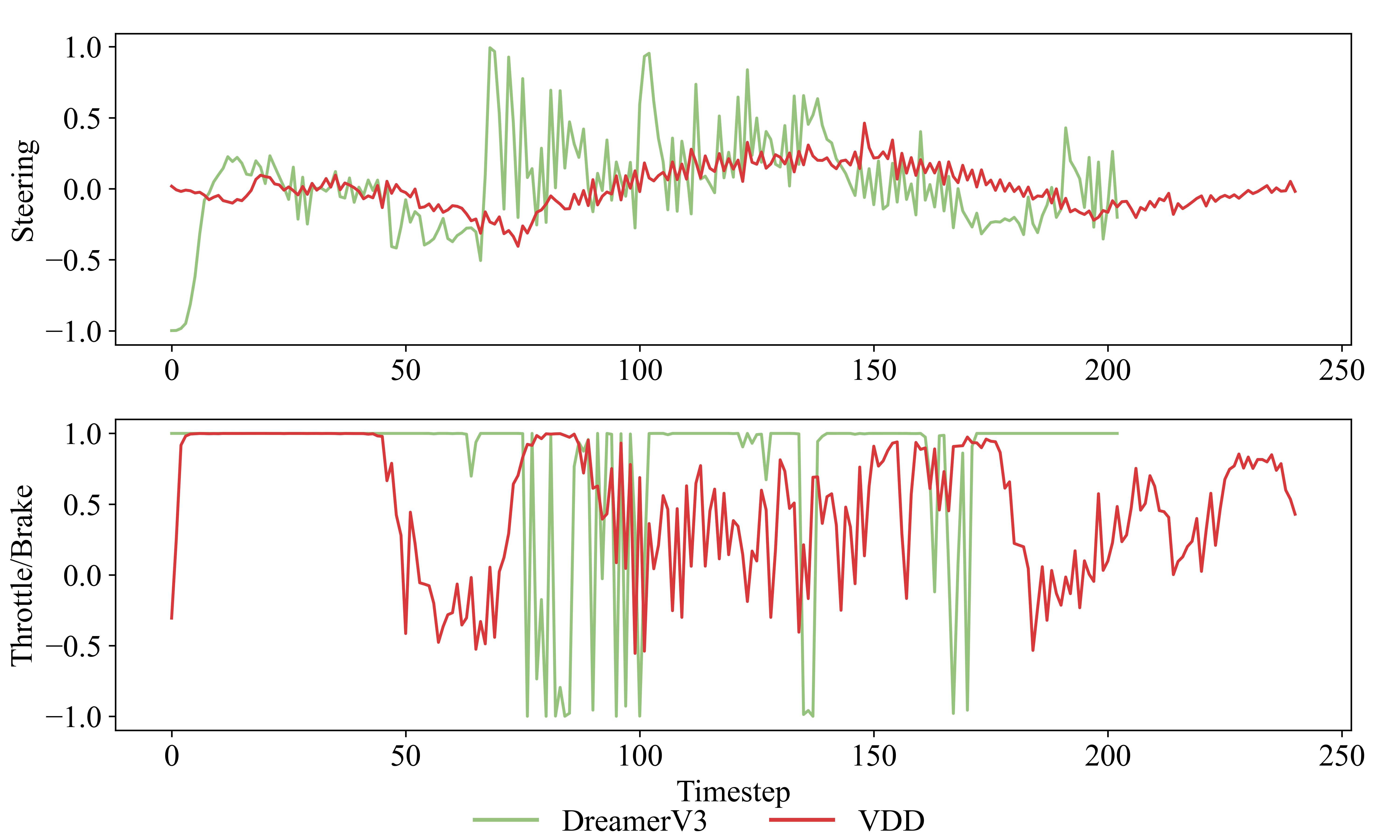}
	\caption{Comparison of steering (top) and throttle/brake (bottom) values for VDD and DreamerV3 in a test episode. }
	\label{fig:action_comparison}
\end{figure}

\begin{figure}
	\centering
	\includegraphics[width=0.45\textwidth]{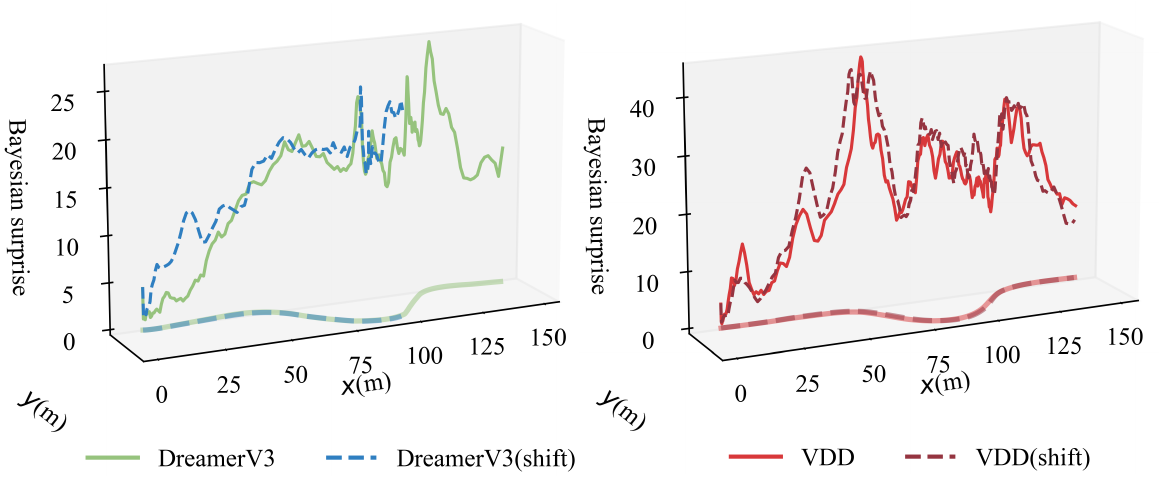}
	\caption{Comparison of Bayesian surprise values in original and shifted environments for DreamerV3 (left) and VDD (right) approaches.}
	\label{fig:surprise}
\end{figure}

To demonstrate the effectiveness of our method, we conducted a case study comparing DreamerV3 and VDD in both original and shifted environments. Both methods were tested using the same random seed for traffic generation under two dynamics parameter scales, (1, 1) and (0.75, 0.75).
Fig.~\ref{fig:traj_comparison} shows the trajectories of the ego vehicle using DreamerV3 and VDD in both environments. Results indicate that in the in-sample environment, both methods reach the destination successfully; however, DreamerV3 exhibits greater action variability, while VDD demonstrates smoother driving performance, as seen more clearly in Fig.~\ref{fig:action_comparison}. In the shifted environment, DreamerV3 fails to stay on the road due to altered vehicle dynamics, resulting in smaller steering angles and increased accelerations when applying actions from the original policy, causing it to deviate from the planned trajectory. Conversely, VDD maintains stable performance despite the dynamics shift, successfully reaching the destination. 
We also plot the Bayesian surprise \cite{itti2009bayesian} for both methods in both environments, which measures the divergence between the prior and posterior distributions. 
The top of Fig.~\ref{fig:surprise} shows the surprise values for DreamerV3 in both environments, with notably higher values in the 0–25m and 75–100m sections, indicating model divergence from the real environment. 
While the VDD method generally shows higher surprise values than DreamerV3, the increase is minor in shifted environments, suggesting that VDD exhibits greater robustness to changes in vehicle dynamics parameters.

\subsubsection{Visualizations}
\begin{figure*}
	\centering
	\begin{minipage}[b]{\textwidth}
        \centering
        \includegraphics[width=1.0\textwidth]{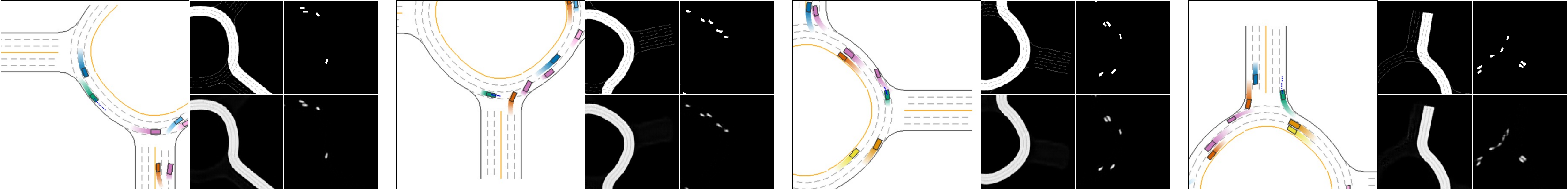}
    \end{minipage}
	\caption{ {Visualization of the trained VDD agent navigating the MetaDrive roundabout maps. In each subplot, the left panel shows the traffic scene at the current moment, dark blue dashed line indicating the ego vehicle's planned trajectory , while the right panel displays the observed BEV image at the top and the reconstructed BEV image at the bottom.}}
	\label{fig:vis-vdd}
\end{figure*}

Visualizations of the proposed VDD agent driving on the intersection and roundabout maps are shown in Fig.~\ref{fig:vis-vdd}. The agent can reconstruct the BEV observation of the surrounding environment precisely and drive in dense traffic safely and efficiently.  {To enhance post-hoc interpretability of the decision outcome, we leverage the world model’s rollout capability to simultaneously visualize the vehicle’s own predicted future trajectory.}

\section{Conclusion}
\label{conclusion}
In conclusion, this paper introduces the Vehicle Dynamics-embedded Dreamer (VDD) algorithm for decision-making in autonomous driving. By decoupling vehicle dynamics modeling from environmental dynamics and employing vehicle parameter-independent dynamic states as inputs for the environmental transition model, the proposed approach significantly enhances generalization across different vehicle platforms. We further propose two strategies to improve policy robustness: Policy Adjustment during Deployment (PAD), which directly adjusts actions based on current and original vehicle parameters during deployment, and Policy Augmentation during Training (PAT), which enhances policy learning through imagined data generated by the world model.
Experimental results in the MetaDrive simulator show that our methods effectively improve driving performance and policy robustness over existing approaches.
 {
While the VDD method achieves strong performance in simulation, we admit that its capability to solve real-world scenarios should be improved. Specific design and more realistic simulators are required to overcome this limitation. In future work, we aim to pretrain the model using offline data, deploy the VDD method on real vehicles and solve potential sim-to-real  and continual learning problems. }. Additionally, we plan to extend this approach to handle unexpected vehicle situations, supporting fault-tolerant decision-making and control to improve the safety and robustness of autonomous driving systems.

\begin{figure}
	\centering
	\includegraphics[width=0.46\textwidth]{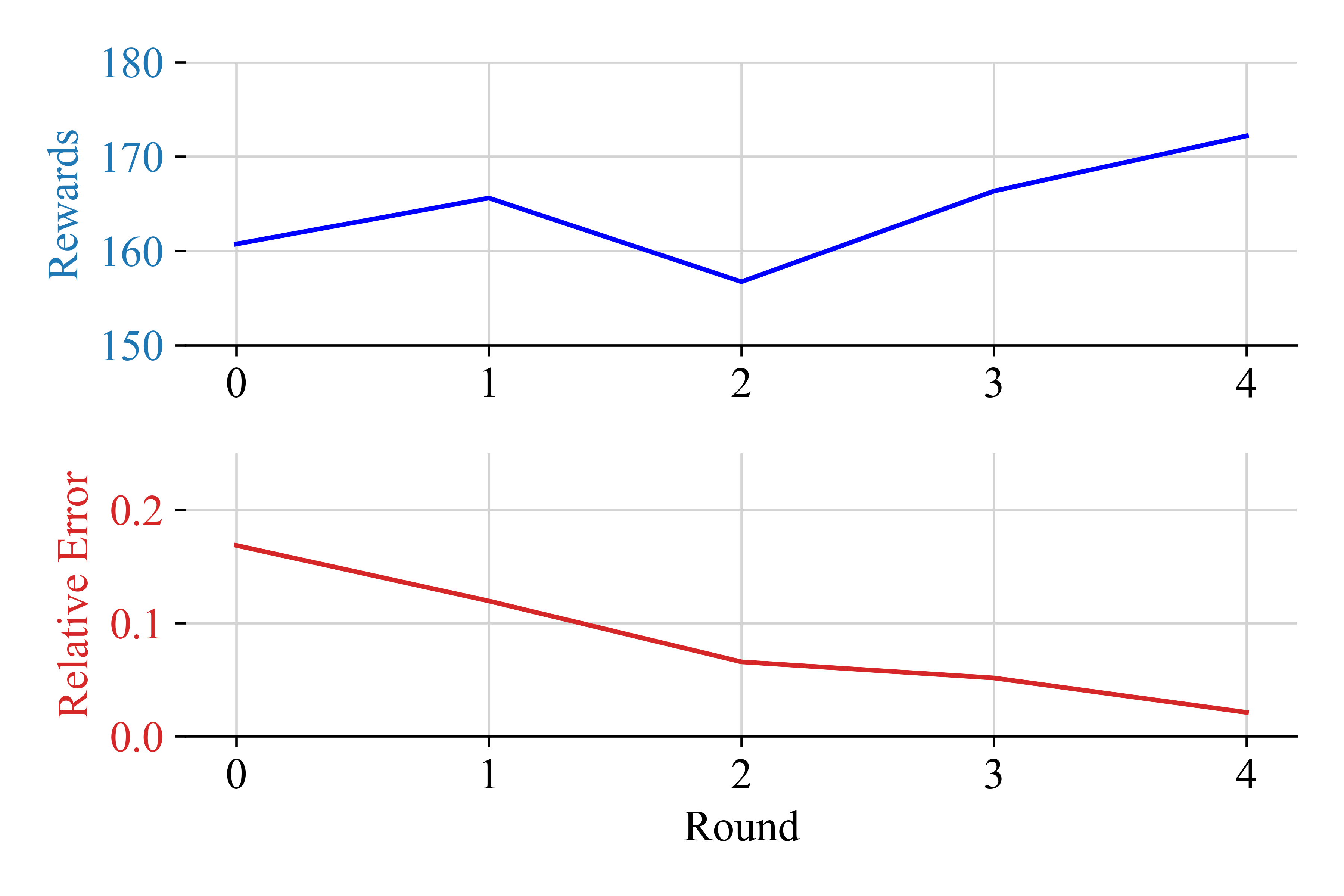}
	\caption{The relationship between episodic rewards or relative error of the vehicle parameters and the number of parameters learning round. }
	\label{fig:dynamics_learning}
\end{figure}

\appendix
\label{appendix}
In the post-deployment phase, we freeze the VDD-PAT policy network and allow only the vehicle-dynamics parameters to be updated. We start from a biased set of vehicle-dynamics parameters, run 10 episodes per round, and record the average return. After each round, the parameters are updated with the collected vehicle-dynamics data before the next round of testing begins. The results are presented in Fig.~\ref{fig:dynamics_learning}. The horizontal axis represents the round number, while the vertical axes show the average episodic rewards and the ratio of the parameter error to the ground truth values. The experimental results demonstrate that online learning effectively updates the vehicle-dynamics parameters, driving them closer to the ground truth values. Moreover, as the error decreases, the cumulative reward per round exhibits an upward trend.

\ifCLASSOPTIONcaptionsoff
  \newpage
\fi

\bibliography{reference.bib}

\begin{IEEEbiography}[{\includegraphics[width=1in,height=1.25in,clip,keepaspectratio]{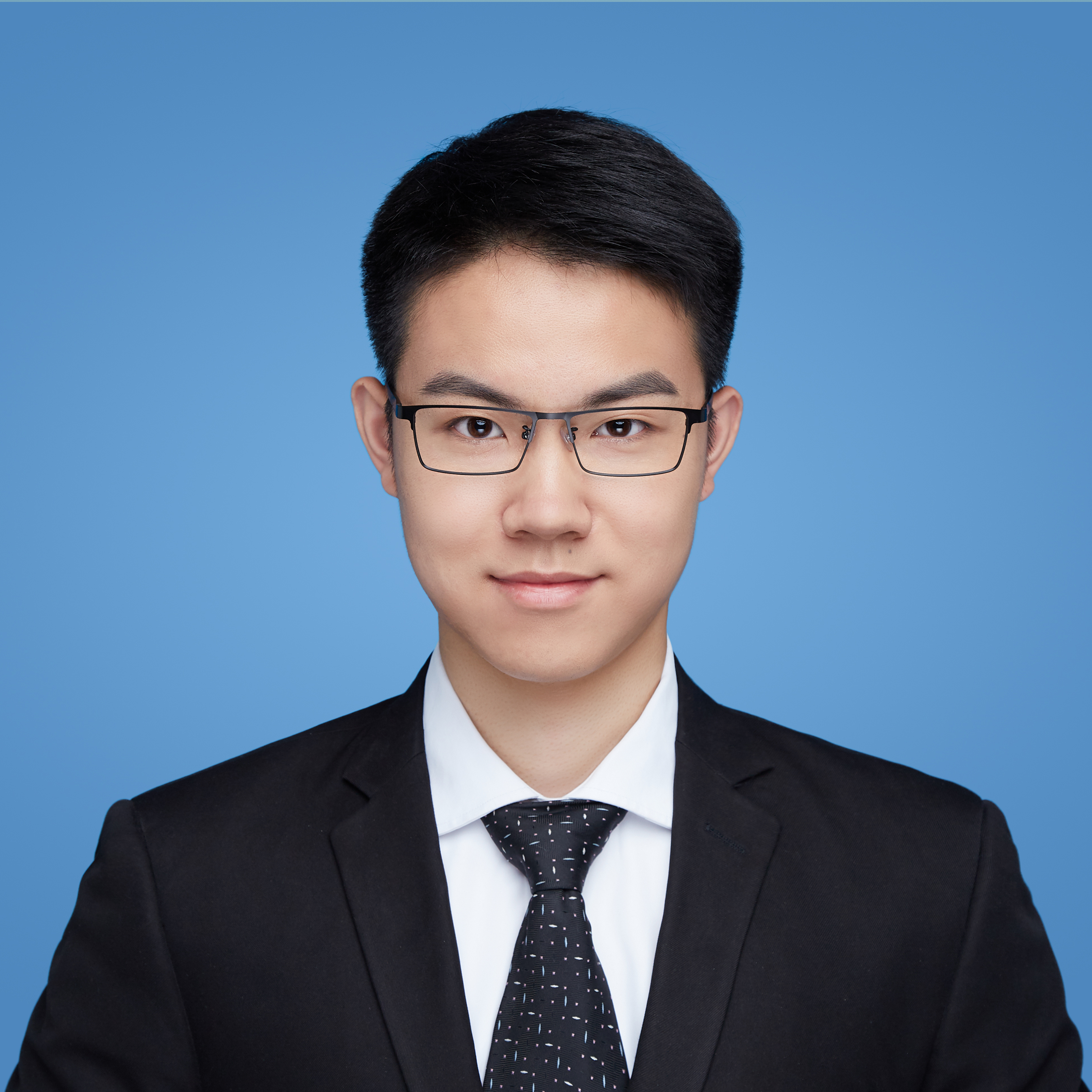}}]{Huiqian Li}
	received the B.E. degree from the
	School of Mechanical Engineering, Beijing Institute of Technology, Beijing,
	China. Currently, he is a Ph.D. student in the School of Vehicle and Mobility in Tsinghua University,
	Beijing, China. He was also a Joint Ph.D. student with the Centre for AI Fundamentals
	and the Centre for Robotics and AI, The University
	of Manchester, U.K. His research fields include world models, reinforcement learning and continual learning in autonomous driving.
\end{IEEEbiography}

\begin{IEEEbiography}[{\includegraphics[width=1in,height=1.25in,clip,keepaspectratio]{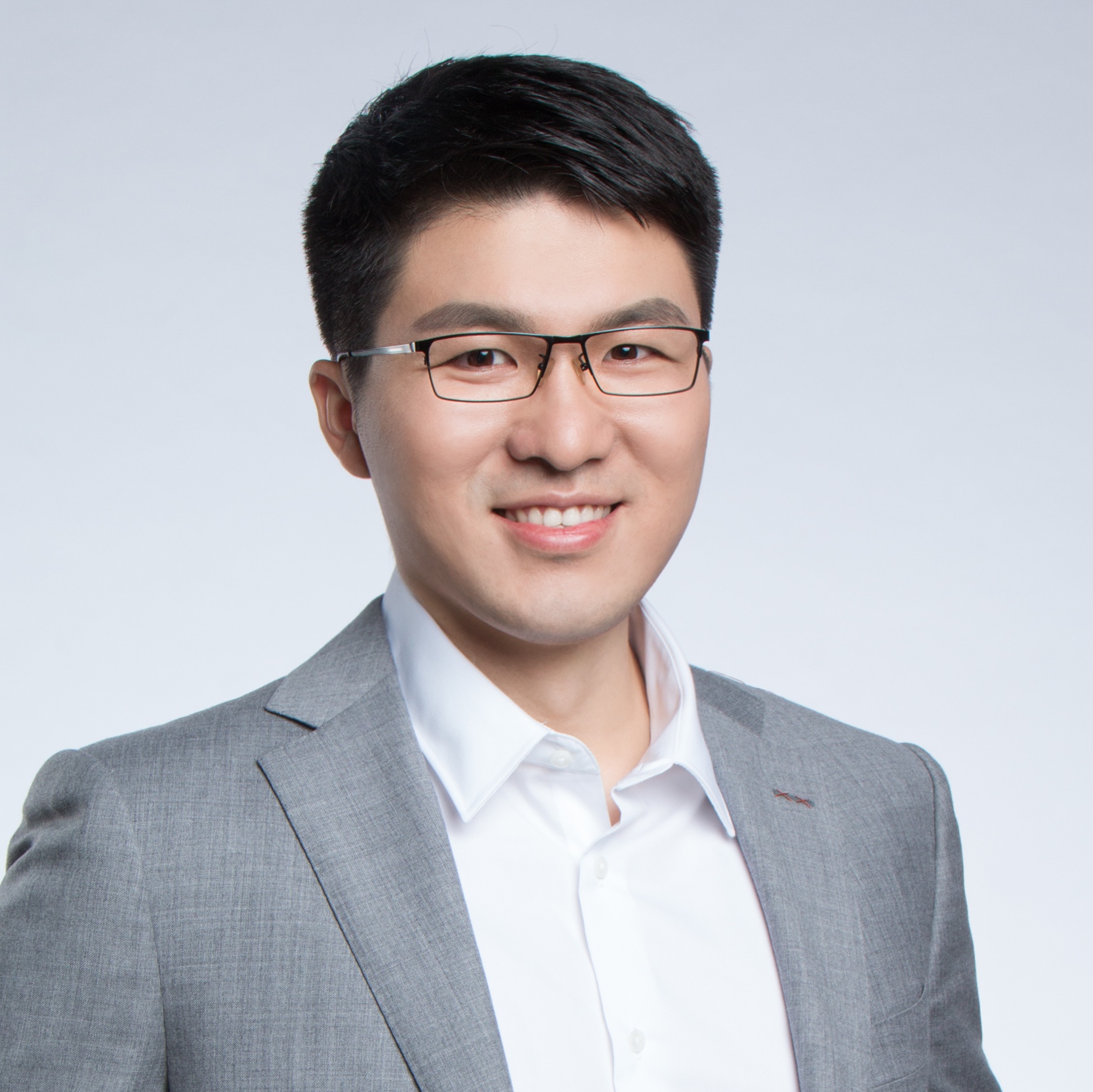}}]{Wei Pan} (Member, IEEE)
	received the Ph.D. degree
	in bioengineering from the Imperial College London,
	London, U.K., in 2016.
	He is currently a Senior Lecturer in machine
	learning with the Department of Computer Science
	and a member of the Centre for AI Fundamentals
	and the Centre for Robotics and AI, The University
	of Manchester, U.K. Before that, he was an Assistant
	Professor in robot dynamics with the Department
	of Cognitive Robotics and the Co-Director of Delft
	SELF AI Laboratory, TU Delft, The Netherlands.
	His research interests include machine learning and control theory with
	applications in robotics.
\end{IEEEbiography}

\begin{IEEEbiography}[{\includegraphics[width=1in,height=1.25in,clip,keepaspectratio]{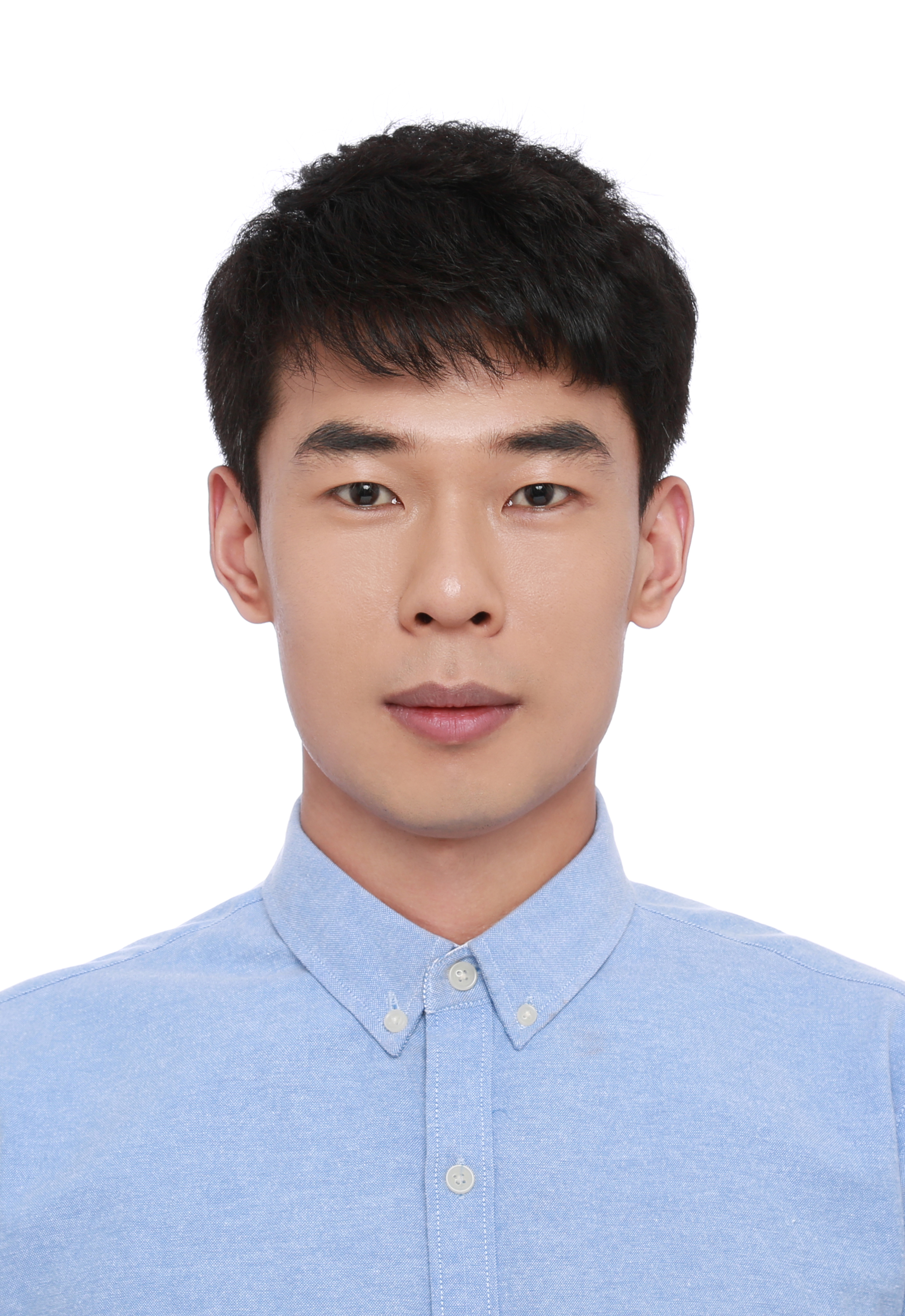}}]{Haodong Zhang} received the B.S. degree in vehicle engineering from Northeast Forestry University, Harbin, China, in 2016, the M.S. degree in vehicle operation engineering from Jilin University, Changchun, China, in 2021, and the Ph.D. degree in Mechanical Engineering in Beijing Institute of Technology, Beijing, China. He is currently a postdoctoral fellow at Tsinghua University. His main research interests are human-vehicle interaction, driver behavior analysis and application of micro-nano sensors.
\end{IEEEbiography}

\begin{IEEEbiography}[{\includegraphics[width=1in,height=1.25in,clip,keepaspectratio]{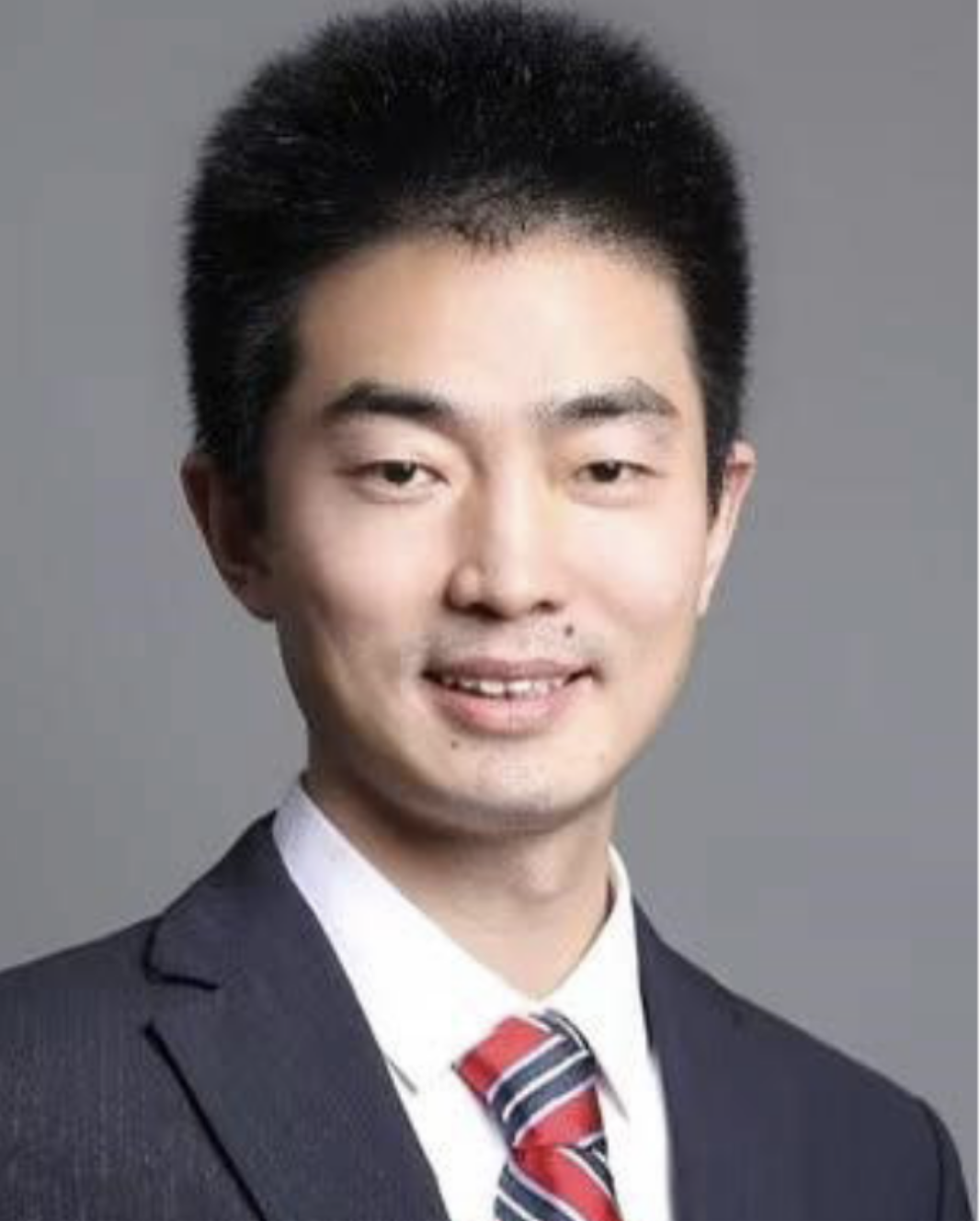}}]{Jin Huang}
received the B.E. and Ph.D. degrees
from the College of Mechanical and Vehicle Engineering, Hunan University, Changsha, China, in
2006 and 2012, respectively. He was a Joint Ph.D.
Student with the George W. Woodruff School of
Mechanical Engineering, Georgia Institute of Technology, Atlanta, GA, USA, from 2009 to 2011.
Then, he started his career as a Post-Doctoral
Fellow and an Assistant Research Professor with
Tsinghua University, Beijing, China, in 2013 and
2016, respectively. His research interests include
artificial intelligence in intelligent transportation systems, dynamics control,
and fuzzy engineering.
\end{IEEEbiography}

\begin{IEEEbiography}[{\includegraphics[width=1in,height=1.25in,clip,keepaspectratio]{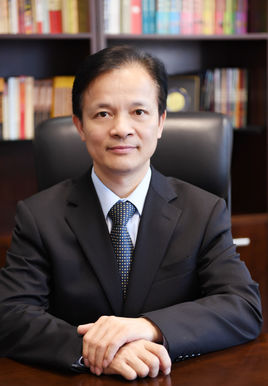}}]{Zhihua Zhong}
received the Ph.D. degree in engineering from Linköping University, Linköping, Sweden,
in 1988. He is currently a Professor with the School of
Vehicle and Mobility, Tsinghua University, Beijing,
China. He was an Elected Member of the Chinese
Academy of Engineering in 2005. His research interests include auto collision security technology, the
punching and shaping technologies of the auto body,
modularity and light-weighing auto technologies, and
vehicle dynamics.
\end{IEEEbiography}

\end{document}